\documentclass[letterpaper, 10 pt, conference]{ieeeconf}  %

\IEEEoverridecommandlockouts                              %

\usepackage{graphicx} %
\usepackage{amsmath}
\usepackage{tikz}
\usepackage{booktabs}
\usepackage{multirow}
\usepackage{makecell}
\usepackage{caption}
\usetikzlibrary{
  arrows.meta,
  positioning,
  fit,
  calc,
  decorations.pathreplacing
}

\usepackage{xspace}

\newcommand{\ie}{\mbox{i.\,e.}\xspace}

\newcommand{\etal}{\emph{et al.}\xspace}
   
\renewcommand{\[}{\begin{equation}}
\renewcommand{\]}{\end{equation}}

\renewcommand{\baselinestretch}{0.99}

\usepackage[capitalize]{cleveref}

\crefname{figure}{Fig.}{Figs.}
\Crefname{figure}{Figure}{Figures}
\crefname{section}{Sec.}{Secs.}
\Crefname{section}{Section}{Sections}
\Crefname{table}{Table}{Tables}
\crefname{table}{Tab.}{Tabs.}
\crefname{algorithm}{Algo.}{Algos.}
\Crefname{algorithm}{Algorithm}{Algorithms}
\crefname{appendix}{Sec.}{Secs.}
\Crefname{appendix}{Section}{Sections}
\usepackage{amsmath}
\usepackage{amssymb}
\usepackage{tensor}

\DeclareMathAlphabet{\mathcal}{OMS}{cmsy}{m}{n}

\def\L{\ensuremath{\mathcal{L}}}
\def\M{\ensuremath{\mathcal{M}}}

\def\P{\ensuremath{\mathcal{P}}}

\def\P{\ensuremath{\mathcal{P}}}

\newcommand{\mat}[1]{\ensuremath{\mathbf{#1}}}

\renewcommand{\det}[1]{\left|#1\right|}
\newcommand{\cardinality}[1]{\left|#1\right|}

\newcommand{\prob}[1]{\ensuremath{p\left(#1\right)}}
\newcommand{\probc}[2]{\ensuremath{\prob{#1 \;\middle\vert\; #2}}}
\newcommand{\probdist}[2]{\ensuremath{p_{#1}\left(#2\right)}}

\newcommand{\set}[1]{\ensuremath{\left\{#1\right\}}}
\newcommand{\fset}[2]{\ensuremath{\set{#1 \;\middle\vert\; #2}}}

\newcommand{\tf}[3]{\tensor[^{#1}]{\mat{#2}}{_{#3}}}

\newcommand{\tuple}[1]{\left\langle #1\right\rangle}

\DeclareMathOperator{\event}{e}

\def\event{e}
\newcommand{\eSDT}[3]{\event^{(#1)}_{#2,#3}}
\newcommand{\eDT}[2]{\event_{#1,#2}}

\newcommand{\ourmethod}{TaPeR}

\title{\LARGE \bf {
TaPeR: Probabilistic Recovery of Sparse Task Precedence Graphs from a Handful of Demonstrations
}}

\author{Adrian Röfer$^{1}$, Karla Stepanova$^{2}$,  and Abhinav Valada$^{1}$%
\thanks{$^1$ The authors are with the Department of Computer Science, University of Freiburg, Germany.}
\thanks{$^2$ Karla Stepanova is with the Czech Institute of Informatics, Robotics, and Cybernetics, Czech Technical University, Czech Republic.}
\thanks{\noindent This work was funded by the BrainLinks-BrainTools center of the University of Freiburg and co-funded by the European Union under the project Robotics and Advanced
Industrial Production (reg. no. CZ.02.01.01/00/22\_008/0004590) and by the Junior STAR project
PersonalRobot no.~26-22610M, funded by the Czech Science Foundation.}
}

\begin{document}
\bstctlcite{IEEEexample:BSTcontrol} %

\maketitle
\thispagestyle{empty}
\pagestyle{empty}

\begin{abstract}
Long-horizon manipulation tasks are often only partially ordered. For example, when assembling an electronic device, the battery and circuit board may be installed in either order, but both must be in place before the enclosure is closed. Recovering such dependencies enables robots to flexibly reorder subtasks while preserving task validity. Existing approaches typically infer task structure from human demonstrations using both temporal and symbolic supervision. However, symbolic predicates require explicit grounding, which is difficult to obtain in realistic settings. In this work, we present an approach for extracting task dependency structures from demonstrations using only simple kinematic graphs and distributions over relative object poses. From these representations, our method estimates pairwise task-step-dependency probabilities and uses them to initialize the edge weights of a precedence graph. We then introduce a filtering pipeline that converts this graph of probability estimates into the final task dependency graph. We evaluate our approach on an existing benchmark and on a new dataset comprising longer tasks with more complex dependencies. We find that our method recovers more accurate task structures from fewer demonstrations than the baselines. Finally, we demonstrate that the inferred graphs can be used to generate multiple valid robotic execution orders for the same task.
\end{abstract}

\section{Introduction}

Imitation learning has become the dominant paradigm for teaching robots new skills. Expert demonstrators collect trajectories completing a desired task, such as making coffee, alongside state observations, often RGB images. A function is then fitted to predict the next action given a state, replicating the demonstrated task~\cite{von2024art}. While initial works learn deterministic state-to-action mappings for pick-and-place tasks~\cite{zhang2018deep}, more recent works capture the diversity of ways to address a single task by viewing action prediction as a denoising process~\cite{yang2025covar,chisari2024learning}, and add language as a conditioning signal to integrate multiple tasks in one model~\cite{black2024pi0}. While this paradigm enables impressive manipulation capabilities, its demand for training data stands in stark contrast to human efficiency: humans need only a handful of sessions, whereas these approaches require tens to hundreds of demonstrations.

\begin{figure}
    \centering
    \includegraphics[width=\columnwidth]{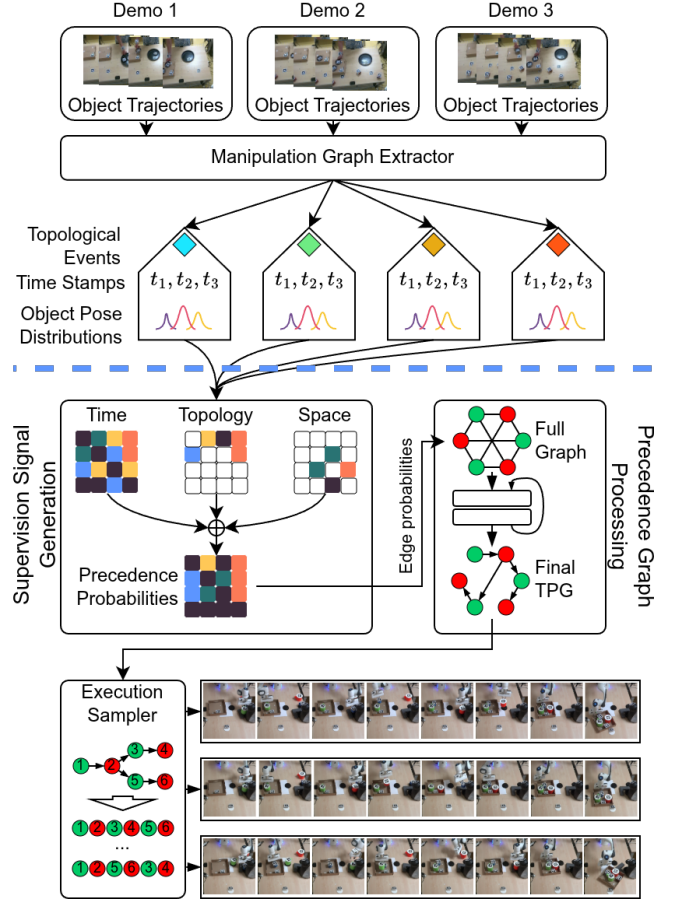}
    \caption{\ourmethod{} identifies independent subtasks in a demonstrated task based on atomic steps identified by an extractor that derives manipulation graphs from demonstrations. \ourmethod{} takes these steps, which represent actions such as \emph{picking up a tray}, and extracts three supervision signals from their grouped observations across multiple demonstrations, which are integrated into precedence probabilities. These serve as edge weights in an initially fully connected graph, which is iteratively reduced to yield a task precedence graph (TPG). From this graph, we can synthesize multiple possible execution sequences of the task and deploy them on a real robot.}
    \label{fig:graphical_problem}
    \vspace{-3mm}
\end{figure}

To boost learning efficiency, researchers reduce model complexity by fitting parametric distributions conditioned on the robot's end-effector pose~\cite{figueroa2018physically}, adopt a real-to-sim approach~\cite{heppert2026scaling}, or use time and low-dimensional signals such as gripper state~\cite{vhartz2026unreasonable}. These works learn meaningful trajectories from no more than a dozen demonstrations, but still require that all of them be executed in the same order, an assumption that becomes difficult as tasks grow in length and is generally unnatural for human demonstrators. Whether one first chops apples or oranges when making a fruit salad is irrelevant, but both must be placed in the bowl before it is mixed and served. Resolving this requires viewing manipulation not as motion but as achieving a change in the world~\cite{kruger2011object}. This view has led to frameworks such as Semantic Event Chains, which model actions as changes in graphical relationships among objects over time~\cite{aksoy2015modelfree}, thereby introducing discrete states whose transition probabilities implicitly encode possible execution orders. Pardowitz \etal~\cite{pardowitz2007incremental} instead explicitly study independent subtasks, identifying the minimal satisfying Task Precedence Graph (TPG) that enforces precedence unless a counterexample is observed. A TPG is a directed acyclic graph (DAG) over task steps whose paths assert
mandatory precedence, leaving unconnected steps independent. Later works build on this, integrating semantic information from action predicates~\cite{ye2019robot}.

In this paper, we contribute to this line of research by simplifying the problem's prerequisites. Following~\cite{aksoy2015modelfree,merlo2025exploiting,rofer2026sparta}, we reduce actions to kinematic changes in object relationships. These changes are singular moments in time, removing the need to explicitly model parallelism in executions. In addition, \ourmethod{} exploits geometric sub-symbolic observations of relative object pose distributions as a proxy for labeled action pre- or post-conditions used in classic semantic supervision. From these, \ourmethod{} derives a graph of precedence probabilities, from which we distill the true precedence graph. Our evaluation shows that these semantic observations enable much earlier identification of task independence than sequence observations alone.
Concretely, our contributions are the following:
\begin{enumerate}
    \item A probabilistic formulation of task step dependence that fuses temporal, topological, and spatial evidence over a handful of demonstrations into pairwise precedence probabilities.
    \item A graph-construction pipeline that turns these pairwise probabilities into a sparse TPG, robust to few, reordered, and noisy demonstrations.
    \item A new dataset of longer and more complex manipulation tasks, together with an evaluation against temporal baselines.
    \item We will release the code and dataset upon acceptance.
\end{enumerate}

\section{Related Work}

Understanding dependence and independence in robotic manipulation tasks was popularized in the context of \emph{Programming by Demonstration} (PbD), though recovering task ordering by observing humans predates it. Early work recognizes tasks from contact-relation transitions~\cite{ikeuchi1995assembly} or from visual recognition of human action sequences~\cite{kuniyoshi1994learning}, synthesizing corresponding robot programs. Zöllner \etal~\cite{zoliner2005towards} introduce a \emph{precedence graph} whose nodes are \emph{manipulation segments}, a complete manipulation from first grasp to final placement, and use different observed execution orders to determine action independence. Subsequent work~\cite{pardowitz2007incremental} realizes this approach and identifies relevant environmental features via human feedback. Later methods add semantic features. Ye \etal~\cite{ye2019robot} introduce semantic effects such as a cucumber transitioning from \emph{whole} to \emph{chopped}, FOON~\cite{paulius2016functional} represents both object states and motions as graph nodes, and Dreher \etal~\cite{dreher2026unified} study the (in)dependence of actions across hands in bimanual tasks, identifying \emph{Allen} relations and, recently, execution timings~\cite{dreher2024learning}.

A second line of work recovers hierarchical rather than flat structure. Hayes and Scassellati~\cite{hayes2016autonomously} construct \emph{Clique/Chain} Hierarchical Task Networks from a graphical task representation, where cliques capture interchangeable subtasks, and chains capture sequential precedence, and CircuitHTN~\cite{chen2021learning} compiles unstructured demonstrations into an HTN via an action graph of valid subsequences. Others learn such structure from large video corpora~\cite{jang2023multimodal,mao2023action}. Box2Flow~\cite{li2024box2flow} recovers an acyclic, directed flow-graph task structure from instructional videos, treating steps with non-overlapping objects as parallel, and Neural Task Graph Networks~\cite{huang2019neural} infer a \emph{conjugate task graph} from a single demonstration. Herbert \etal~\cite{herbert2026learning} recover bimanual structure with a GNN encoder and transformer decoder.

We employ a TPG task model as in~\cite{pardowitz2007incremental,ye2019robot}, but our nodes are not manipulation segments. They are changes in the scene's kinematic graph~\cite{aksoy2011learning,merlo2025exploiting,rofer2026sparta}, identifying a single moment rather than an interval, which removes the need for interval algebra over actions \cite{dreher2024learning,dreher2026unified} and lets precedence be estimated pairwise. We capture these changes through distributions of relative object poses which, together with kinematic-graph membership, form our only supervision, requiring no classifier for categorical spatial relations~\cite{pardowitz2007incremental,paulius2016functional,ye2019robot}. Unlike clique/chain and action-graph approaches~\cite{hayes2016autonomously,chen2021learning} that read structure off a largely clean task graph, we construct the graph probabilistically from a few reordered, noisy demonstrations, explicitly resolving contradictory orderings. Compared to learning-based models~\cite{huang2019neural,jang2023multimodal,herbert2026learning} that \emph{predict} future actions from substantially more data, we infer \emph{hard} precedence and independence constraints from a handful of demonstrations. Like Box2Flow~\cite{li2024box2flow}, we use object overlap as the primary dependence cue, but obtain it probabilistically from trajectories rather than a learned detector over annotated video, and recover a general TPG rather than a tree.

\section{Problem Formulation}
\label{sec:problem_formulation}

We assume we are given a linear sequence of atomic steps $s_1,\ldots,s_N$ that represent a task, extracted from a set of $\mathcal{D}$ demonstrations. The aim is to connect these linear steps in a directed acyclic graph (DAG) in which the existence of a path $s_i \leadsto s_j$ expresses that step $s_i$ must always precede $s_j$ in any possible execution of this task. This model of task variations has been coined \emph{task precedence graph} (TPG) by Pardowitz~\etal~\cite{pardowitz2007incremental}.
Different from previous models, each step $s_n$ represents a transition in the control state of an object $o$, marking either the activation (acquiring control) or deactivation (releasing control) of that object. A step collects observations of the same kinematic change across the set of demonstrations $\mathcal{D}$ and collects the timestamps at which it was observed.
A step $s_n$ is always associated with two sets of objects: the predicted objects $\P_n$ and the landmarks $\L_n$, which serve as static reference objects during the execution of the step. The step captures a distribution $\prob{\tf{\L_n}{T}{\P_n}}$ over the relative poses of the objects of $o \in \P_n$ with respect to objects $o' \in \L_n$. %
In addition, there is a set of manipulators $\M$.
If $\M \cap \L_n \not = \emptyset$, then $s_n$ is referred to as an \emph{activation}-step, otherwise, $s_n$ is a \emph{deactivation}-step. This reflects the assumption that actions are executed by obtaining control of an object (activating it) and subsequently releasing it (deactivating it). 
We denote the sets of activation and deactivation steps by $\mathcal{S}^+$ and $\mathcal{S}^-$, respectively.

Given the extracted steps and their associated pose distributions, our objective is to recover a \emph{partially ordered} task abstraction that explains permissible temporal variation and spatial dependencies while preserving manipulation semantics, as illustrated in \cref{fig:graphical_problem}. %

\section{Method}

In the following, we introduce our method, which aggregates three distinct but partially overlapping supervision signals to identify dependence and independence among observed task steps. Fundamental is the observed temporal order ($T_{ij}$), which is then supported by object interaction dependencies ($D_{ij}$) and spatial configuration dependencies between object placements ($S_{ij}$) to make a proposal of a path probability $\prob{s_i \leadsto s_j}$ estimate. These path probabilities are then used to guide the incremental pruning of an initially fully connected precedence graph.
Throughout, we denote hyperparameters using the letter $\alpha$.

\subsection{Trajectory Preprocessing and Step Extraction}

The atomic steps $s_n$ used in our method are extracted from object trajectory demonstrations following the procedure in~\cite{rofer2026sparta}. We briefly summarize the process here.

First, kinematic connections between objects are identified over time, yielding a sequence of graphs $G_t$. Each graph $G_t$ is decomposed into connected subgraphs $H_{m,t}$ for each manipulator $m \in \M$, and an additional residual graph $H_{W,t} = G_t - \bigcup_{m\in \M} H_{m,t}$. 
An \emph{event} is detected whenever the cardinality of a connected component increases, \ie, $\cardinality{H_{*,t}} > \cardinality{H_{*,t-1}}$.
For such an event, we define the predicted objects $\P_\event = H_{*,t} \setminus H_{*,t-1}$ and landmark objects $\L_\event = H_{*,t} - \P_\event$. Since objects are consistently identified across demonstrations, events with similar topology can be matched across different demonstrations. 
We identify the collection of topologically compatible events across demonstrations:
\[
    E_n = \left(\eSDT{n}{1}{t_1}, \ldots, \eSDT{n}{\cardinality{\mathcal{D}}}{t_{\cardinality{\mathcal{D}}}}\right),
\]
where ($d,t_d$) indicate the demonstration and time at which the event was observed. 
Note that not every event has to be identified in every demonstration.
The events in $E_n$ are organized such that the first event $\eSDT{n}{1}{t_1}$ serves as a \emph{root event}. Candidate events in other demonstrations, denoted by $\eSDT{n}{>1}{*}$, are selected based on their topological compatibility with the root under the matching relation $\simeq$. Consequently, multiple candidates may satisfy this condition within a demonstration. This induces a set of possible cross-demonstration matches:
\[
    \hat{\mathcal{E}_n} = \set{\eSDT{n}{1}{t_1}} \times \ldots \times \fset{\eDT{\cardinality{\mathcal{D}}}{t}}{\eDT{\cardinality{\mathcal{D}}}{t} \simeq \eSDT{n}{1}{t_1}}.
\]

Each step $s_n$ is therefore associated with both the event collection $E_n$ and the set of possible matches $\hat{\mathcal{E}}_n$. 
For each event, pose observations of objects $o \in \P_\event$ relative to $o' \in \L_\event$ are available as $\tf{W}{T}{o,t_d}^{(d)}$. These observations are used to estimate the step-level pose distributions.

\subsection{Temporal Dependency}

The observed temporal order provides the first source of evidence for precedence relationships between individual steps (we refer to this as a temporal supervision signal). While prior work~\cite{pardowitz2007incremental} employs temporal supervision as a hard discriminator and~\cite{dreher2024learning} uses ordinal counting with a confusion zone, we propose a soft temporal prior that grows in confidence with the number of observations. For each pair of steps ($s_i$,$s_j$), let $\mathcal{D}'$ denote the set of demonstrations in which both steps co-occur. For each demonstration $d \in \mathcal{D}'$, we record whether $s_i$ was observed before $s_j$. Under the null hypothesis of no directional temporal preference, both orders ($i\leadsto j$ and $j\leadsto i$) are equally likely. We thus model the observed ordering as samples from a binomial distribution and estimate the posterior probability that $s_i$ precedes $s_j$. Let $T_{ij}=\probc{s_i \leadsto s_j}{{\mathcal{D}'}}$ denote the observed posterior of the precedence relation $s_i \leadsto s_j$, estimated from the set of demonstrations $\mathcal{D}'$ in which both steps occur. We use a beta conjugate prior to derive $T_{ij}$:
\[
\begin{aligned}
    T_{ij} = 1 - I_{\frac{1}{2}}(&1 + \cardinality{\set{s_i\leadsto s_j \in \mathcal{D}'}}, \\
        &1 + \cardinality{\mathcal{D}'} - \cardinality{\set{s_i\leadsto s_j \in \mathcal{D}'}}),
\end{aligned}
\]
where $I$ denotes the regularized incomplete beta function.

\subsection{Topological Dependency}

Given our assumption of a task consisting of \emph{activations} and subsequent \emph{deactivations} of objects, we can impose a rule that requires all the deactivated objects to have been activated before. 
Formally, for every deactivation step $s_j \in \mathcal{S}^-$, we assume there exists a set of activation steps $\mathcal{A} \subseteq \mathcal{S}^+$ such that the predicted objects $\P_j$ of $s_j$ are contained in the union of the predicted objects of the steps in $\mathcal{A}$, \ie, $\forall s_j \in \mathcal{S}^- \;\exists \mathcal{A} \subseteq \mathcal{S}^+: \P_j \subseteq \bigcup_{s_i\in \mathcal{A}} \P_i$. Through tool use, \ie using a broom to sweep objects into a bin, we can establish a dependence between two $s_i, s_j\in \mathcal{S}^+$ if $\L_j \cap \P_i \not = \emptyset$.

To incorporate these priors at the pairwise level, we introduce the dependency probability $D_{ij} = \probc{s_i \leadsto s_j}{\mathcal{S},\mathcal{P},\mathcal{L}}$ expressing that $s_i$ should precede $s_j$ due to task dynamics.  
As a demonstrator may activate (\ie, pick up and gain control of) multiple objects sequentially and deactivate (\ie, place and release control of) them jointly (or vice versa), we do not require $\P_i = \P_j$. Instead, we consider two steps structurally related if they manipulate at least one common object, \ie, if $\P_i \cap \P_j \neq \emptyset$.
This consideration yields the sparse topological dependency posterior for path-precedence:
\[\small
    D_{ij} = \begin{cases}
        T_{ij} & \text{if}\ s_i \in \mathcal{S}^+ \wedge s_j \in \mathcal{S}^{-} \wedge \P_i \cap \P_j \neq \emptyset \\
        T_{ij} & \text{if}\ s_i,s_j \in \mathcal{S}^+ \wedge (\L_j \cap \P_i) \not = \emptyset  \\
        \alpha_{D} T_{ij}  & \text{otherwise}
    \end{cases},
\]
where $\alpha_D \in (0,1)$ is a background prior.
Effectively, this definition acts as a regularizer on the observed temporal precedence posterior $T_{ij}$ and encodes the prior assumption that activation steps usually precede the deactivation steps because they enable them, while still allowing uncertainty when no object overlap is observed. %

\subsection{Spatial Dependency}

While topological dependence ties deactivation steps from $\mathcal{S}^-$ to the activation steps from $\mathcal{S}^+$, spatial dependence connects $s_i, s_j \in \mathcal{S}^-$ through overlap between the predicted objects of $s_i$ and the landmark objects of $s_j$. The step $s_i$ might need to precede $s_j$ because the former places an object that the latter critically depends on, \ie, the placement of a plate informs where the silverware should be placed.
This dependence is predicated on $s_i$ affecting a landmark of $s_j$ as $\P_i \cap \L_j \neq \emptyset$. Different from topological dependence, the probability estimate $S_{ij} = \probc{s_i \leadsto s_j}{\mathcal{L}_j^*, \mathcal{P}_i}$ does not act as a simple binary switch on $T_{ij}$, but is
weighted by the 
relevance of objects $o' \in \P_i$ for predicting $\tf{\L_j}{T}{\P_j}$, \ie
\[
S_{ij} =
\begin{cases}
T_{ij} \cdot \probc{\L_j^*}{\P_i}, & \text{if } \P_i \cap \L_j \neq \emptyset, \\
\alpha_S \, T_{ij}, & \text{otherwise},
\end{cases}
\]
where $\probc{\L_j^*}{\P_i}$ denotes the estimated probability that the object set $\P_i$ is relevant for predicting the distribution $\prob{\tf{\L_j}{T}{\P_j}}$, and $\alpha_S \in (0,1)$ is again a background prior when no spatial dependency is observed (\ie no manipulated object of $s_i$ serves as a landmark for step $s_j$).

To estimate the relative pose distribution $\prob{\tf{\L_j}{T}{\P_j}}$, we consult the set of cross-demonstration event matches
$\hat{\mathcal{E}_n}$ associated with step $s_j$. 
For each possible cross-demonstration match $E' \in \hat{\mathcal{E}}_n$,  we compute the empirical distribution of relative poses between predicted objects $o \in \P_j$ and landmarks $l \in \L_j$. This yields covariance matrices $\Sigma_{l,o,E'}$ for each object–landmark pair. 
We interpret tight relative-pose distributions (\ie, low dispersion) as evidence that a landmark is structurally relevant for predicting $\tf{\L_j}{T}{\P_j}$. To quantify this, we evaluate the dispersion via the determinant of the covariance matrix $\det{\Sigma_{l,o,E'}}$, where lower values indicate tighter spatial coupling. 
Aggregating over all matches $E' \in \hat{\mathcal{E}}_n$, we compute
\[
\begin{aligned}
\mu_{\Sigma} &= \frac{1}{Z} \sum_{l\in\L_j}\sum_{o\in\P_j}\sum_{E' \in \hat{\mathcal{E}}_n} \det{\Sigma_{l,o,E'}}, \\
\sigma_{\Sigma}^2 &= \frac{1}{Z} \sum_{l\in\L_j}\sum_{o\in\P_j}\sum_{E' \in \hat{\mathcal{E}}_n} 
\left(\det{\Sigma_{l,o,E'}} - \mu_{\Sigma}\right)^2,
\end{aligned}
\]
where $Z$ is a normalizing constant, representing the total number of landmark-object match combinations.

We then assign a relevance score to each landmark $l \in \L_j$ by comparing its average dispersion to the global mean:

\[
\prob{l \in \L_j^*}
=
\frac{
1 - \exp\!\left(
-\frac{\left[\max(\mu_{\Sigma} - \bar{\Sigma}_l,\,0)\right]^2}{2\sigma_{\Sigma}^2}
\right)
}{
\max\limits_{l' \in \L_j}
\left(
1 - \exp\!\left(
-\frac{\left[\max(\mu_{\Sigma} - \bar{\Sigma}_{l'},\,0)\right]^2}{2\sigma_{\Sigma}^2}
\right)
\right)
},
\]
where
\[
\bar{\Sigma}_l = \frac{1}{|\P_j|\,|\hat{E}_n|} 
\sum_{o\in\P_j}\sum_{E' \in \hat{E}_n} \det{\Sigma_{l,o,E'}}.
\]

We normalize by the highest probability value to ensure that at least one landmark attains maximal relevance ($\cardinality{\L_j^*} \geq 1$).
To judge the relevance of the predicted object set $\P_i$, we intersect $\P_i$ and $\L_j$ and contrast this intersection with $\L_j \setminus \P_i$. Both sets' objects are sorted in descending order by $\prob{l\in\L_j^*}$, and the balance is measured by the number of objects in the same position. We pad $\L_j \setminus \P_i$ with $0$ probability if $\cardinality{\L_j} < \cardinality{\P_i}$.

\subsection{Forming a Task Graph}
\label{sec:task_graph}

\begin{figure}
    \vspace{1.5mm}
    \centering
    \includegraphics[height=18mm]{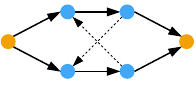}
    \hspace{5mm}
    \includegraphics[height=18mm]{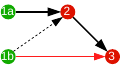}
    \caption{Graphical Depiction of the \emph{stochastic cycle removal} and \emph{ordered contrastive pruning}. \emph{Left}: The graph contains a cycle, with its SCC highlighted in blue, edge thickness indicates weight. Cutting any of the edges between the blue nodes would resolve the cycle. The diagonal edges are equally weak. Our algorithm removes both of them instead of arbitrarily cutting just one. \emph{Right}: The nodes are traversed by their depth in the DAG, indicated by their number in the figure. As the edge $(1b, 2)$ is weaker than $(1a, 2)$, it is removed. This removal blocks the subsequent evaluation of $(1b, 3)$, without re-evaluation of the successor probabilities for $1b$.}
    \label{fig:graph-tricks}
\end{figure}

Our aim is to obtain a partially ordered model of a task as a Directed Acyclic Graph (DAG) without any transitive redundancies. The probability estimates from the previous sections serve as indicators for the order of this graph. As a first step, we integrate them into a combined estimate $C_{ij}$ by overlaying the individual probabilities in the following hierarchy:
\[
\label{eq:path-conflict}
    C_{ij} = \begin{cases}
        D_{ij} & \text{if}\ s_i \in \mathcal{S}^+ \wedge s_j \in \mathcal{S}^- \\
        D_{ij} & \text{if}\ s_i, s_j \in \mathcal{S}^+ \\
        S_{ij} & \text{if}\ s_i \in \mathcal{S}^- %
    \end{cases}.
\]

\textbf{Resolving Path Conflicts}: Given different orders of execution, we will make contradictory observations of execution order: In one demonstration, the pot might get picked up first, while in another it might be the lid. The result is that both steps have a lower, but relevant, probability of being the other's predecessor. We resolve contradictions by performing a contrastive normalization on opposing probabilities:
\[
    C^*_{ij} = C_{ij}\left(\frac{2C_{ij}}{C_{ij} + C_{ji}}-1\right)^2,
\]
which reduces $C^*_{ij}$ to $0$ if $C_{ij} = C_{ji}$.

\textbf{Deriving Edge-Probabilities}: Our observations do not imply an \emph{immediate} precedence of steps, but merely if $s_i$ should occur some time before $s_j$. 
In order to form a graph, we need to convert these \emph{path-precedences} into probability estimates $\prob{s_i \rightarrow s_j}$ for direct edges. We do so, exploiting the following rationale: \emph{if $s_i$ is an immediate predecessor of $s_j$, then there should exist no $s_k$ in between them}. We express this rationale as
\[\label{eq:counter-prob}\small
    \prob{s_i \rightarrow s_j} = C^*_{ij} \left(1 - \max(p^*(i,k,j), p^*(i,k,l,j))\right),
\]
where $p^*(i,...,j)$ denotes the highest probability path of length $2$ or $3$ to $s_j$ starting from $s_i$. This is essentially an approximate marginalization over precedence probabilities across alternative graphs. $p^*(i,...,j)$ is computed as the product of $C^*_{ik}\cdot \ldots \cdot C^*_{*j}$, assuming conditional independence among edges along a path for simplification.
We include paths of length $3$ due to the nature of our signals: Topological supervision generally separates all steps in $\mathcal{S}^+$ from one another, while spatial does the same for most steps in $\mathcal{S}^-$. This soft bi-partition makes paths of length $2$ uninformative.

Using $\prob{s_i \rightarrow s_j}$, we construct a first, fully connected graph $G_0 = (V, E_0)$. In pursuit of the desired DAG, we employ an iterative processing pipeline, which prunes edges until the graph structure anneals. The pipeline consists of five iterative steps executed until the topology stabilizes, followed by one final reduction step.

\noindent\textbf{1) Removing low probabilities}: The first stage is a simple removal of all edges with $\prob{s_i \rightarrow s_j} < \alpha_p$. \\
\textbf{2) Stochastic cycle removal}: We introduce a uniqueness-aware cycle cutting method to transform the graph into a DAG.
First, we obtain all strongly connected components (SCCs) of $G$. While there exists a path $i\leadsto j$ for each pair of nodes in an SCC, not each edge is equally important to the SCC. Given the subgraph $H$ which is an SCC of $G$, we consider the \emph{centrality} $c_B(e)$ of an edge $e=(i,j)$, which is the fraction of shortest paths in $H$ passing through $e$. We rank edges $e \in H$ by $r(e) = \prob{e}^{-1} c_B(e)$ in descending order and remove edges until $H$ is acyclic. This reveals the number $N_c$ of edges needed to cut in order to resolve the cyclicity of $H$. However, multiple edges in the cycle can have similar probabilities and be equally capable of dissolving the cycle, so they should all be removed. In order to find these, we sample other sets of edges $\mathcal{C}_i$ of size $N_c$ according to the weights $r(e)$: $\mathcal{C} = \{\mathcal{C}_1,\ldots,\mathcal{C}_{\cardinality{C}}\}$. We score each successful set $\mathcal{C}_i$ as $r(\mathcal{C}_i) = 1 - \prod_{e\in \mathcal{C}_i} \prob{e}$ and measure the \emph{decisiveness} $d(\mathcal{C}_i)$ of a candidate by contrasting it with all others:
\[
    d(\mathcal{C}_i) = \max_{\mathcal{C}_m \in \mathcal{C}\backslash{\mathcal{C}_i}} \frac{r(\mathcal{C}_i)}{r(\mathcal{C}_m)}.
\]
Afterwards, we normalize $d(\mathcal{C}_i)$ by the $\max(d(\mathcal{C}_i))$ and then cut all $\mathcal{C}_i$ with $\hat{d}(\mathcal{C}_i) > \alpha_\mathcal{C}$.\\
\textbf{3) Ordered contrastive pruning}: Diverging probabilities of incoming edges are an indicator of the graph's instability, which should be removed. Let $\mathcal{I}(j)$ be the set of incoming edges for $s_j$, then we normalize $\prob{e_i}$ by $\max_{e\in \mathcal{I}(j)}\prob{{e}}$ and remove $e_i$ if $\hat{p}(e_i) < \alpha_O$.
Notably, we apply this filter in the topological order of the graph: Passing from the roots to the leaves. At every node, we mark all nodes $s_i$, if we remove an edge $e = (i,j)$. By removing the edge, we change how \cref{eq:counter-prob} would be computed. This invalidates the nominal value of the remaining edges of $i$, thus no further pruning can be done on $i$ without re-calculating the edge weights.\\
\textbf{4) Annealing the topology}: The previous contrastive pruning can only process parts of the graph every iteration, as pruning dirties the weights of the remaining edges.
We thus iterate the pipeline, setting the path probability $p(i\leadsto j) = 0$ if there is no path $i\leadsto j \in G_1$. Given this new bias, we again compute $\prob{s_i \rightarrow s_j}$, build the graph, and filter it as before. We repeat this process until the nodes' adjacency in the graph no longer changes.\\
\textbf{5) Limited transitive reduction}: The final graph $G^*$ might still contain transitive redundancies, such as a direct connection, $i\rightarrow j$, and an indirect one $i\rightarrow k \rightarrow j$. 
A standard transitive reduction does not account for the weight of the edges along alternative paths. We propose the following weight-aware version: Given the weakest probability of an edge along a path
\[
    \probdist{\min}{p = \tuple{n_1,\ldots,n_M}} = \min_{m<M} \prob{s_m\rightarrow s_{m+1}},
\]
we find $\probdist{\max\min}{i,j} = \max\fset{\probdist{\min}{p}}{p = i \leadsto j}$, and only remove a redundant connection $e = (i, j)$ if $\prob{e} \leq \probdist{\max\min}{i,j}$.
This final step concludes our graph processing pipeline, yielding the task's stable TPG.

\section{Experimental Evaluation}
\label{sec:eval}

\begin{figure}
    \vspace{1.5mm}
    \centering
    \includegraphics[width=0.32\columnwidth]{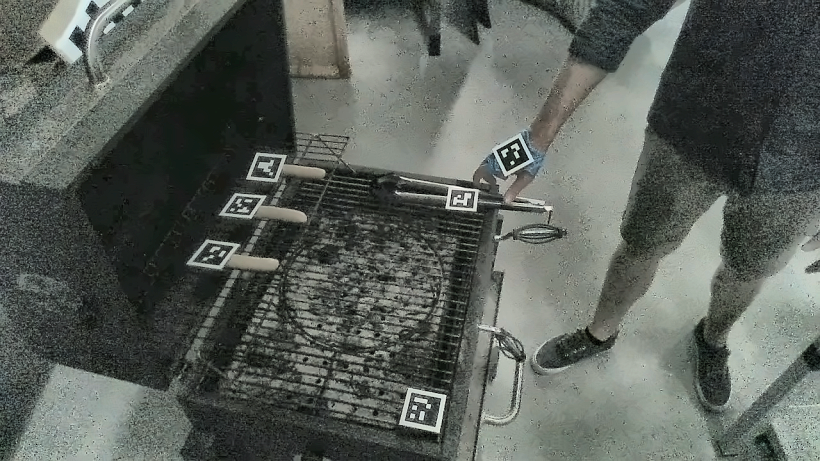}
    \includegraphics[width=0.32\columnwidth]{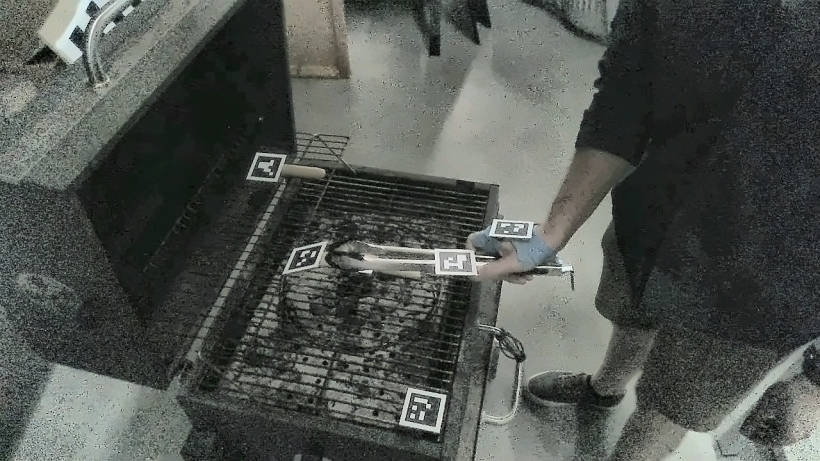}
    \includegraphics[width=0.32\columnwidth]{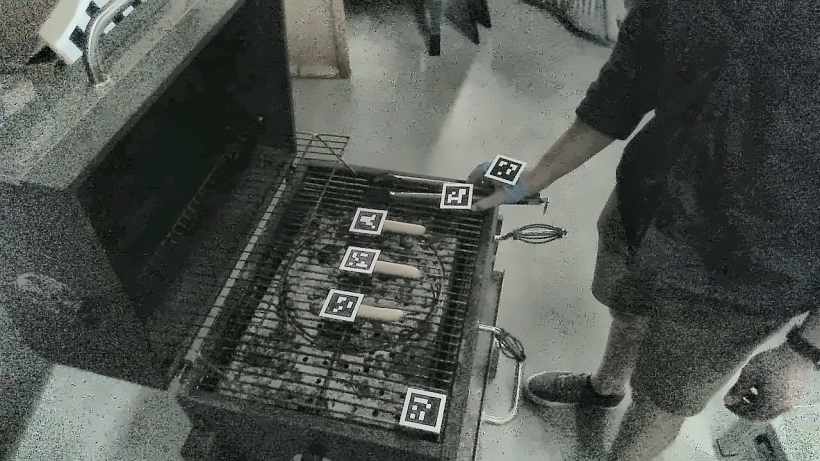}
    \\ \vspace{1mm}
    \includegraphics[trim={200 100 300 100},clip,width=0.32\columnwidth]{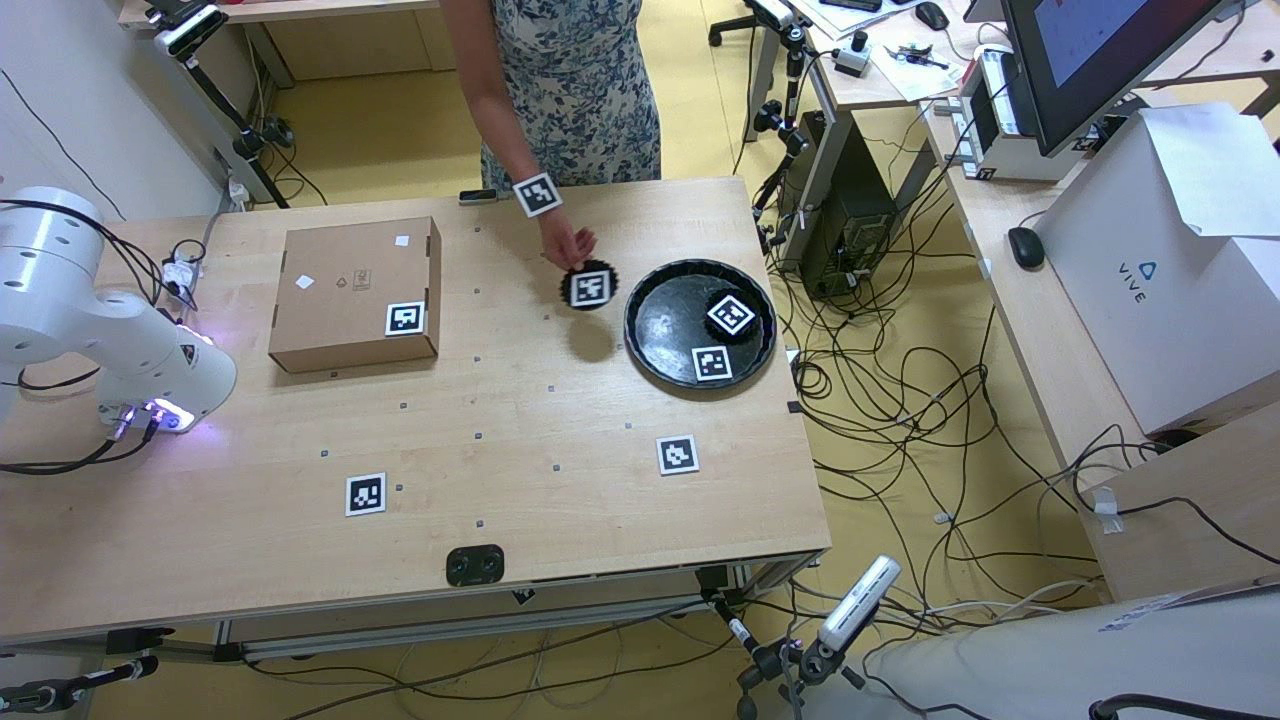}
    \includegraphics[trim={200 100 300 100},clip,width=0.32\columnwidth]{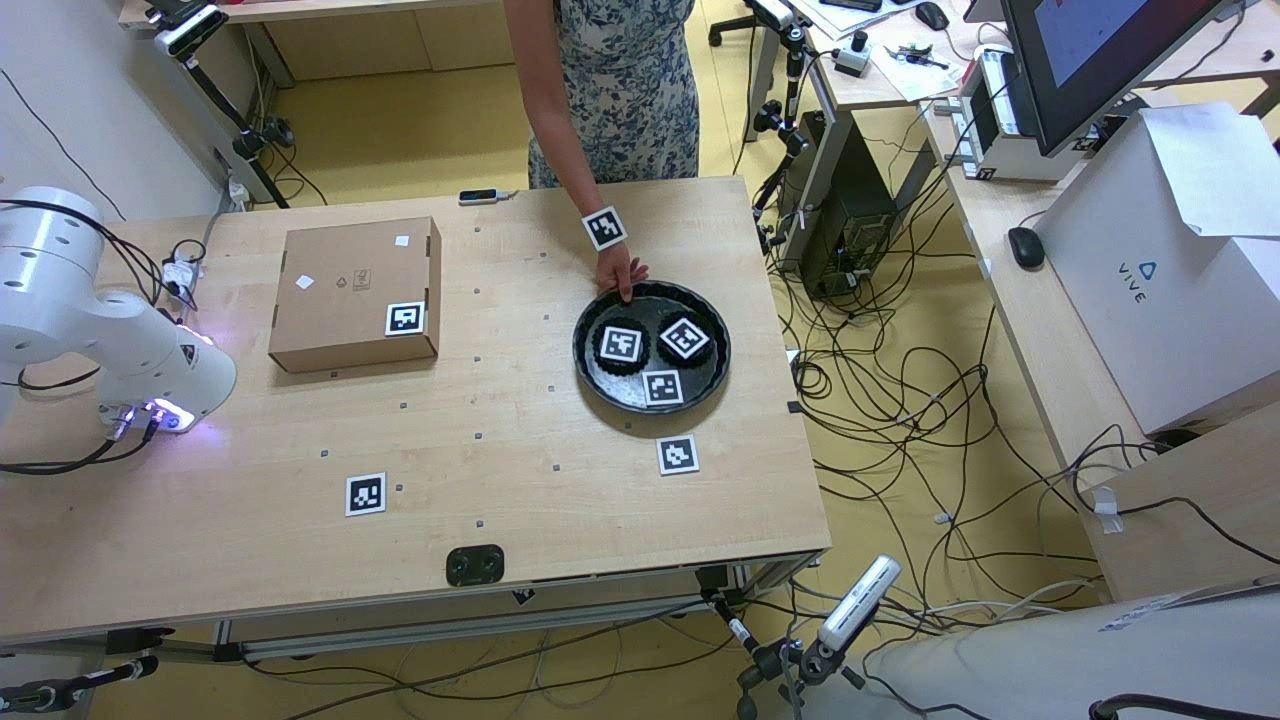}
    \includegraphics[trim={200 100 300 100},clip,width=0.32\columnwidth]{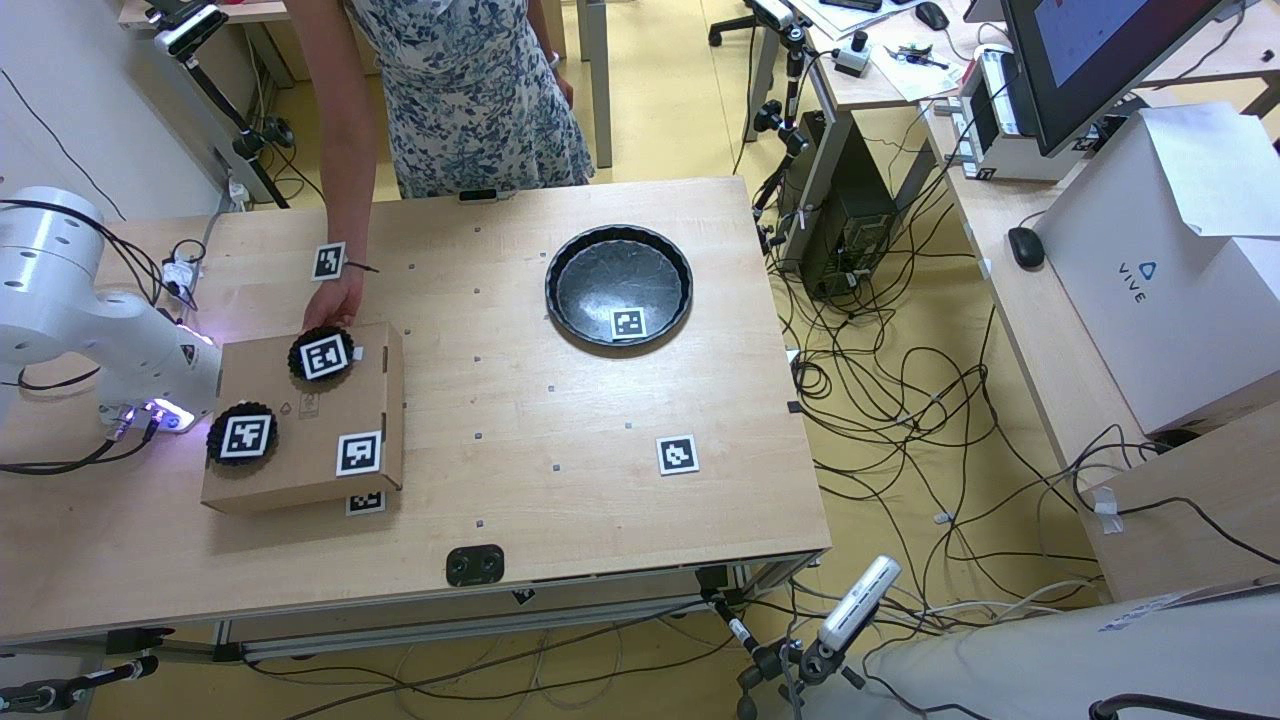}
    \\ \vspace{1mm}
    \includegraphics[trim={200 100 300 100},clip,width=0.32\columnwidth]{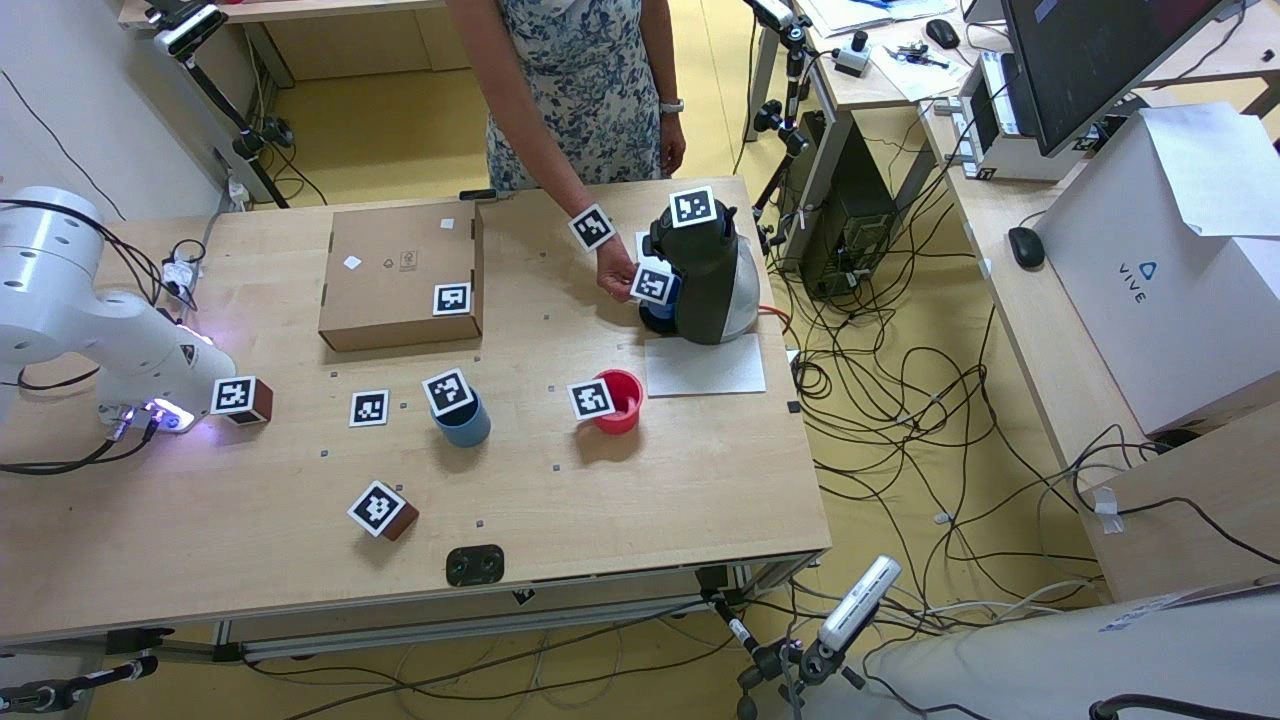}
    \includegraphics[trim={200 100 300 100},clip,width=0.32\columnwidth]{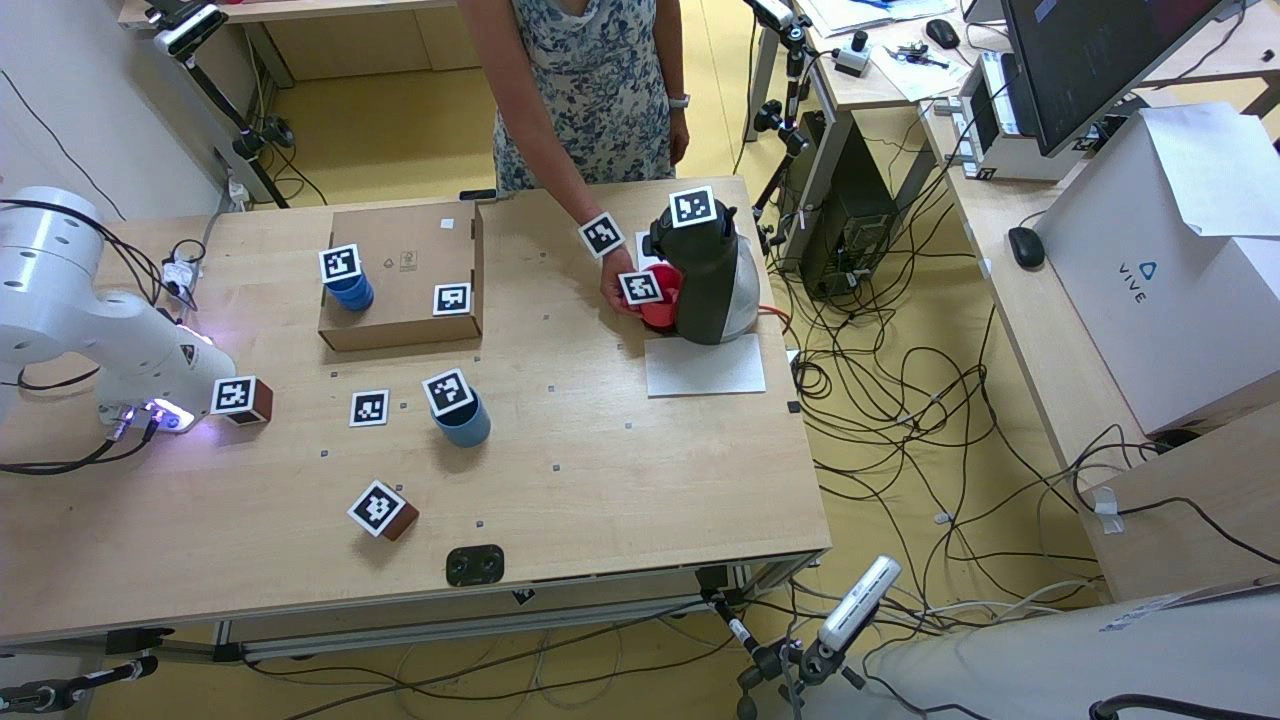}
    \includegraphics[trim={200 100 300 100},clip,width=0.32\columnwidth]{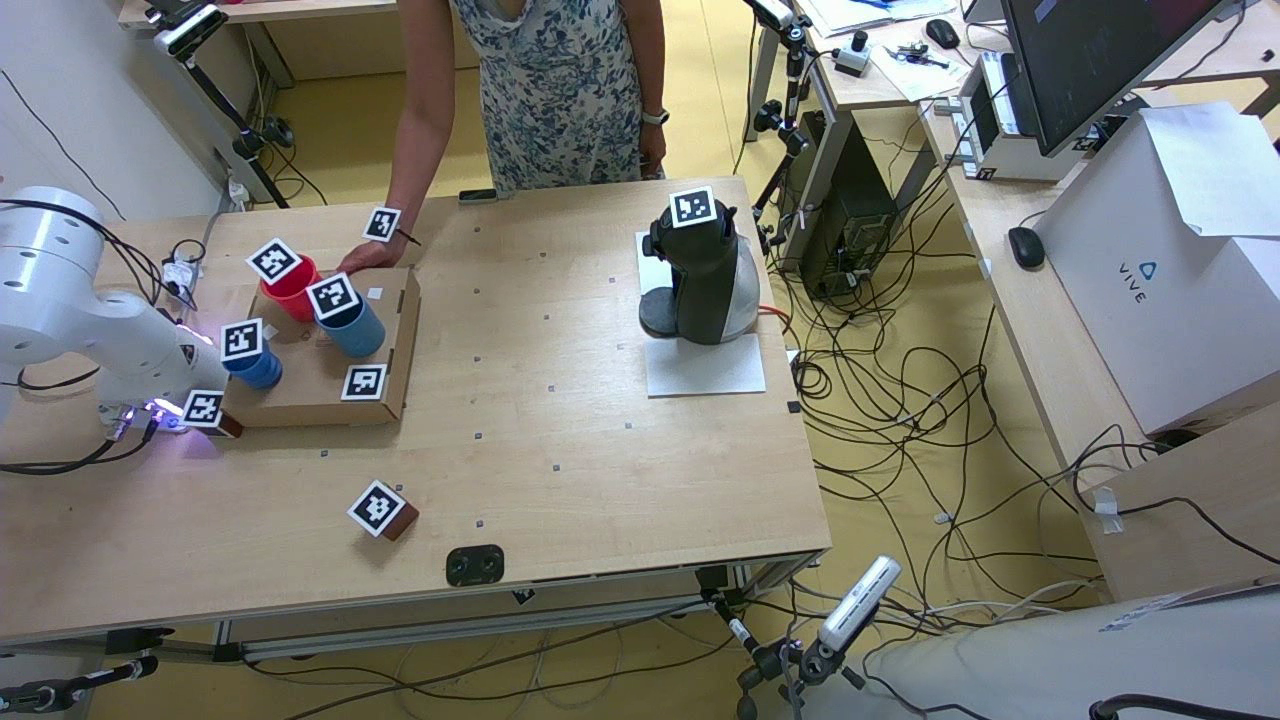}
    \caption{Qualitative impression of HANDSOME-COMPLEX tasks. \textit{Top:} Moving sausages onto the lower grill using tongs. Picking up the tongs is a prerequisite for manipulating the sausages. \textit{Middle:} Making and delivering muffins. The muffin cups are moved to a baking tray, which is then placed in an oven. After removal from the oven, the muffin cups are moved to a tray, which is moved to a delivery location. \textit{Bottom:} Making three coffees and delivering them on a tray. The order of coffee making is independent, but the tray's movement depends on all three cups being made.}
    \label{fig:cvut_data}
    \vspace{-3mm}
\end{figure}

We evaluate the benefit of our semantic supervision compared with typical temporal supervision. Additionally, we examine the utility of the individual supervision signals and measure the impact of different stages in our graph processing pipeline to assess the value of each component. Throughout the evaluation we use $\alpha_D=\alpha_S=0.05$, $\alpha_p=0.2$, $\alpha_\mathcal{C}=0.95$, and $\alpha_O=0.8$.

\ourmethod{} requires observations of 3D object poses, which is a restriction not shared by prior works~\cite{pardowitz2007incremental,dreher2024learning}.
The HANDSOME dataset~\cite{merlo2025exploiting} satisfies the requirement. It contains 16 uni-manual or bi-manual tasks, each with 25 demonstrations. However, these are largely linear or independent and rather short, with none exceeding 4 steps. To address this lack of complexity, we introduce 8 new tasks, each a longer sequence of up to 14 steps, with an average of $9.6$ steps per task.
Aside from longer sequences of pick-and-place tasks, our demonstrations also include transportation tasks in which sausages are moved using tongs, or cups are loaded onto trays and moved using these. We present a qualitative impression of our dataset in \cref{fig:cvut_data}.
We create a ground-truth task graph for each task against which we compare the inferred instances.
We generate task models by uniformly sampling demonstrations for a particular task with sample sizes $(2-7)$, extract events from them following~\cite{rofer2026sparta}, and match them within the sampled pools. 
Our primary quantitative metric is the F1-score for edge existence in a graph. Given the graph's sparsity, we find other metrics to be heavily biased towards false negatives.

\begin{figure*}[]
    \centering
    \vspace{3mm}
    \includegraphics[width=0.9\textwidth]{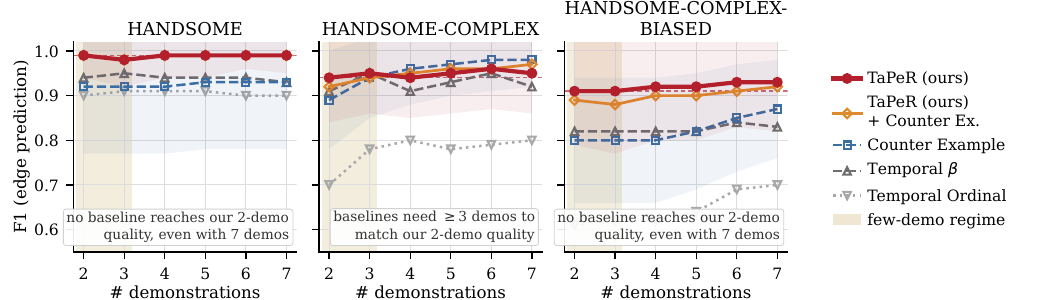}
    \caption{We compare our method against pure temporal supervision across the HANDSOME~\cite{merlo2025exploiting} dataset and our HANDSOME-COMPLEX (HC) extension. We uniformly sample 2-7 demonstrations for a task and use their matched events for our method, pure temporal observation, and the counterexample-based method proposed in~\cite{pardowitz2007incremental}. We report the F1-score over edge predictions in the final TPG. We find that our Semantic supervision performs best in the settings with few demonstrations, where temporal evidence is sparse. While the counterexample becomes the strongest in HC, we note that this is not the case in HANDSOME. We can reproduce the same pattern by sampling demos in HC with a preference towards one particular execution order. Semantic supervision can ignore this bias, while all temporal methods decline. Across all datasets and demo counts, we find our $\beta$-prior to be better than an ordinal temporal signal. We use~\cite{pardowitz2007incremental} as temporal prior for its late model improvement, but find the effect to be slightly negative.}
    \label{tab:results-1}
    \vspace{-3mm}
\end{figure*}

\subsection{Comparison with Temporal Supervision}
We compare the final model quality given the maximum number of training demonstrations ($7$) to assess whether semantic supervision is advantageous given ample training data. In addition, we compare our binomial temporal prior to an ordinal one, as used in \cite{dreher2024learning}, and the counterexample-based connectivity proposed by~\cite{pardowitz2007incremental}. To make the graphs more similar to our expected forms, we apply a transitive reduction to retain only necessary edges.
We also study two types of data deficiency: 1) Human task execution bias, 2) Faulty event matches.  In collecting HANDSOME-COMPLEX, we ensured sufficient randomization of execution orders to support temporal observations.
For studying human execution bias, we sample this dataset in a biased manner, favoring similar execution orders.
For the study of event mismatches, we introduce a probability that structurally similar events within a demonstration will be confused for one another, causing them to trade places. 

\textbf{Results:} On HANDSOME-COMPLEX, our semantic supervision performs best when deployed on smaller pools of training samples $(2-3)$. Afterward, the counter-example-based heuristic introduced by Pardowitz~\etal~\cite{pardowitz2007incremental} yields both higher scores and lower variance. Interestingly, this is not the case for the simpler HANDSOME dataset. We investigate the disparity and note that HANDSOME does not provide sufficient execution variety in its demonstrations for some tasks, making it impossible or unlikely to observe alternative executions.
When comparing temporal methods, we can see that ordinal temporal supervision is very brittle, overcommitting to imbalanced observations. The $\beta$-prior performs significantly better, especially on few data samples.
Given these findings, we also ablate \emph{\ourmethod{} + Counter Example}, which replaces the temporal prior with the simple binary flags proposed by~\cite{pardowitz2007incremental}. We find that our approach gains minimally at best, while losing significantly in the absence of temporally diverse demonstrations.
In our study of execution bias, we observe that it is detrimental to all temporal methods, whereas the performance of semantic supervision decreases only slightly.
We report the impact of event mismatching separately in \cref{tab:r1_noise}. Also in this experiment, we observe deterioration in both semantic supervision and the baseline~\cite{pardowitz2007incremental}, but by a wide margin. The aggressive pruning of connections by~\cite{pardowitz2007incremental} becomes detrimental when events are mismatched.
Notably, the effect is smallest on the original HANDSOME dataset. This is due to the demonstrations being rather short and containing almost no repeated manipulation of the same object, resulting in few structurally equivalent events to confuse.

\begin{table}[]
    \centering
    \vspace{2mm}
    \begin{tabular}{ll|ccccc}
    \toprule
    & & \multicolumn{5}{c}{Event Mismatch Chance} \\
    Dataset & Config & $0\%$ & $5\%$ & $10\%$ & $20\%$ & $30\%$ \\
    \midrule
    \multirow{3}{*}{\makecell[l]{HANDSOME- \\ COMPLEX}} & \ourmethod{} & \textbf{0.95} & \textbf{-0.01} & \textbf{-0.03} & \textbf{-0.05} & \textbf{-0.07} \\
     & \makecell[l]{Counter- \\ Example} & \textbf{0.95} & -0.02 & -0.07 & -0.15 & -0.20 \\
    \cmidrule(lr){2-7}
    \multirow{2}{*}{\makecell[l]{HANDSOME- \\ COMPLEX-\\BIASED}} & \ourmethod{} & \textbf{0.92} & \textbf{-0.00} & \textbf{-0.02} & \textbf{-0.04} & \textbf{-0.06} \\
     & \makecell[l]{Counter- \\ Example} & 0.82 & -0.01 & -0.04 & -0.10 & -0.14 \\
    \cmidrule(lr){2-7}
    \multirow{3}{*}{HANDSOME} & \ourmethod{} & \textbf{0.99} & \textbf{-0.00} & \textbf{-0.00} & \textbf{-0.00} & \textbf{-0.01} \\
     & \makecell[l]{Counter- \\ Example} & 0.92 & \textbf{-0.00} & -0.01 & -0.01 & -0.02 \\
    \bottomrule
    \end{tabular}
    \caption{We measure the impact of events being matched wrongly across demonstrations. We report the F1-score for edge predictions with a perfect event match and the delta from this baseline across different noise levels. While both our method and the counterexample-based supervision from~\cite{pardowitz2007incremental} are negatively impacted by matching noise, we observe a wide gap ($\approx2x$) between the two methods.}
    \label{tab:r1_noise}
   \vspace{-3mm}
\end{table}

\subsection{Analysis of Method}

\textbf{Impact of supervision signals:} In pursuit of understanding the contributions of the components of our method, we start by isolating the supervision signals. We use the binomial temporal prior $T_{ij}$ and decay it by either $D_{ij}$ or $S_{ij}$. We measure the change in F1-Score as a fraction of the score produced by $T_{ij}$. We report the results of this experiment in \cref{tab:r2_results}, alongside the improvement provided by the combined supervision. The result is counterintuitive: the individual supervision signals are often destructive.
In our HANDSOME-COMPLEX dataset, they always reduce the model score individually. In the biased case, they can outperform temporal observation half the time, while they mostly improve the model extracted from the HANDSOME data.
Despite being at best mediocre on their own, their combined use always yields a significant improvement.
This behavior is caused by $D_{ij}$ and $S_{ij}$ affecting two separate sets of edges. $D_{ij}$ decays edges between steps in $\mathcal{S}^+$ and edges $\mathcal{S}^+ \rightarrow \mathcal{S}^-$, while $S_{ij}$ scores edges $\mathcal{S}^-$ and edges $\mathcal{S}^- \rightarrow \mathcal{S}^+$. Thus, running either supervision individually biases the graph towards the unscored edge set. When temporal supervision is poor, as in HANDSOME and our biased sampling of HANDSOME-COMPLEX, this is likely to yield an improvement as the graph is probably over-connected. If temporal supervision is strong, it biases the graph's structure.

\begin{table}[]
    \centering
    \setlength{\tabcolsep}{4pt}
\begin{tabular}{ll|rrrr}
\toprule
Dataset & Config & \multicolumn{1}{c}{2} & \multicolumn{1}{c}{3} & \multicolumn{1}{c}{5} & \multicolumn{1}{c}{7} \\
\midrule
\multirow{3}{*}{\makecell[l]{HANDSOME- \\ COMPLEX}} & Topological ($D_{ij}$) & -3.6\% & -3.7\% & -2.6\% & -1.7\% \\
 & Spatial ($S_{ij}$) & -3.8\% & -3.4\% & -2.5\% & -1.4\% \\
 & Combined & \textbf{4.3}\% & \textbf{0.8}\% & \textbf{2.5}\% & \textbf{3.5}\% \\
\cmidrule(lr){2-6}
\multirow{3}{*}{\makecell[l]{HANDSOME- \\ COMPLEX- \\ BIASED}} & Topological ($D_{ij}$) & -1.5\% & 1.0\% & 0.9\% & -0.9\% \\
 & Spatial ($S_{ij}$) & -1.5\% & -0.8\% & 2.6\% & 3.0\% \\
 & Combined & \textbf{11.3}\% & \textbf{10.9}\% & \textbf{14.0}\% & \textbf{13.3}\% \\
\cmidrule(lr){2-6}
\multirow{3}{*}{HANDSOME} & Topological ($D_{ij}$) & 1.3\% & 0.4\% & 1.4\% & 2.1\% \\
 & Spatial ($S_{ij}$) & 1.7\% & -0.3\% & 2.3\% & 3.5\% \\
 & Combined & \textbf{5.5}\% & \textbf{4.7}\% & \textbf{6.3}\% & \textbf{8.1}\% \\
\bottomrule
\end{tabular}
    \caption{Relative improvement of individual \ourmethod{}'s supervision signals over temporal signal. Values are reported as percentage change in F1-score produced by the temporal baseline. The combined signal always yields an improvement, whereas the individual signals reduce model quality.}
    \label{tab:r2_results}
    \vspace{-3mm}
\end{table}

\begin{table}[]
    \centering
    \setlength{\tabcolsep}{4pt}
    \begin{tabular}{lrrr}
    \toprule
            & \multicolumn{3}{c}{Dataset}  \\
    Pipeline Stages & H-C & H-CB & H \\
    \midrule
    \makecell[l]{Low probabilities\\\hspace{2mm}+Cycle removal} & 0.81 & 0.76 & 0.99 \\
    \cmidrule{2-4}
    \hspace{2mm}+Contrastive Pruning & +0.10 & +0.11 & +0.00 \\
    \cmidrule{2-4}
    \hspace{5mm}+Transitive Reduction & +0.13 & 0.15 & +0.00 \\
    \cmidrule{2-4}
    \hspace{5mm}+Annealing & -0.02 & -0.00 & -0.01 \\
    \cmidrule{2-4}
    \hspace{5mm}\makecell[l]{+Annealing\\+Transitive Reduction} & +0.14 & 0.16 & +0.00 \\
    \bottomrule
    \end{tabular}
    \caption{We study the impact of our different pipeline stages on model quality. We report the F1-scores for the first two stages required to obtain a DAG from the initial model. For each consecutive stage, we report its delta to the previous one. }
    \label{tab:r3_results}
    \vspace{-3mm}
\end{table}

\textbf{Impact of graph processing stages:} We study the graph processing stages by measuring the graph quality in the form of the F1-score after different stages. As a baseline, we use the first two stages, which remove low-probability edges and apply our soft cycle removal to produce a DAG that we can score. We then append the further stages to this base and report the relative change in \cref{tab:r3_results}. 

Unlike in the study of supervision signals, we find that each processing stage yields approximately equal improvements in graph quality. Notable is the impact of the annealing stage: when applied on its own, it reliably degrades quality, even below the initial two stages of the pipeline, with very low variance. When followed by the transitive reduction, however, it yields the highest performance by a small margin, with the lowest variance. 
We compare cases where the annealing stage yields an improvement over directly applying transitive reduction. It only yields a benefit when graphs remain highly connected after the initial soft cycle removal, so that the contrastive pruning stage cannot reduce any additional connections in a single iteration.

\subsection{Robotic Validation}

\begin{figure}
    \centering
    \includegraphics[width=0.18\columnwidth]{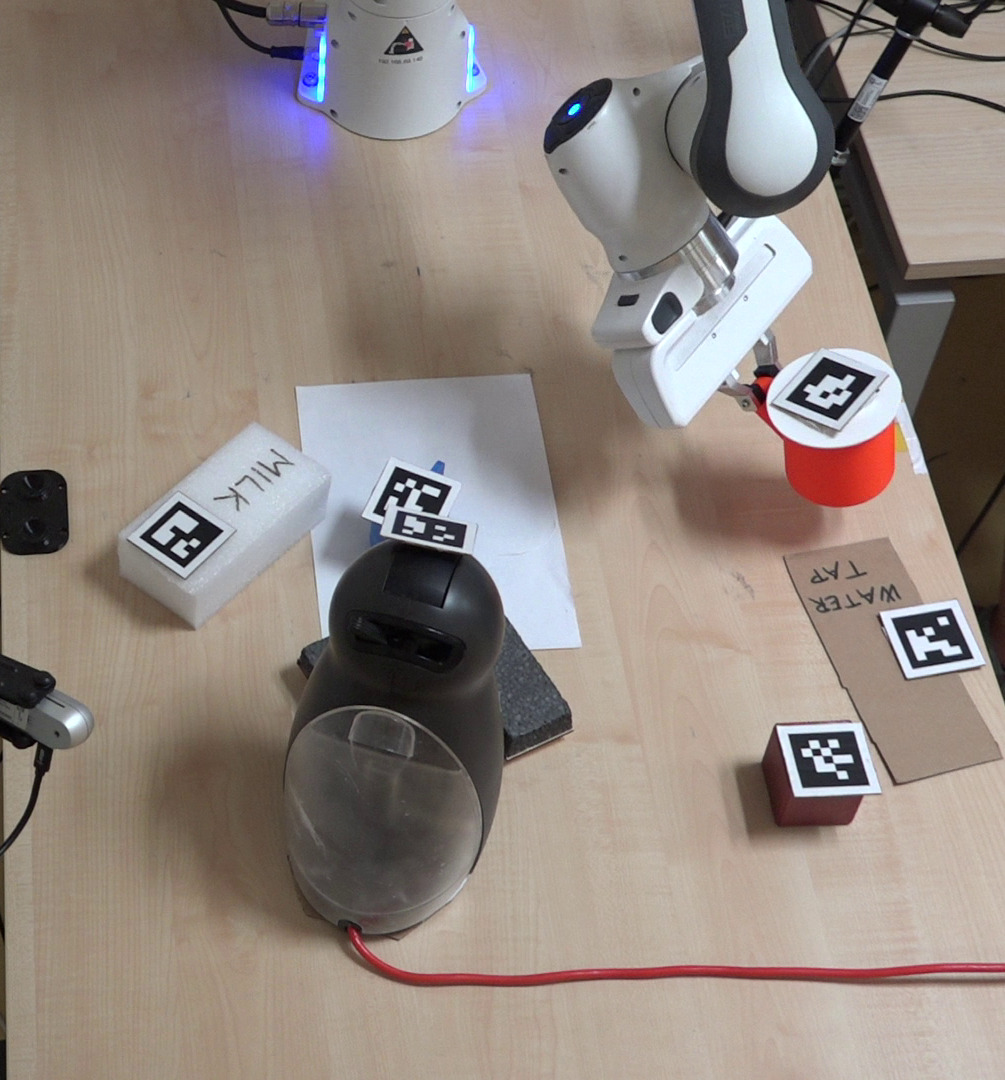}
    \includegraphics[width=0.18\columnwidth]{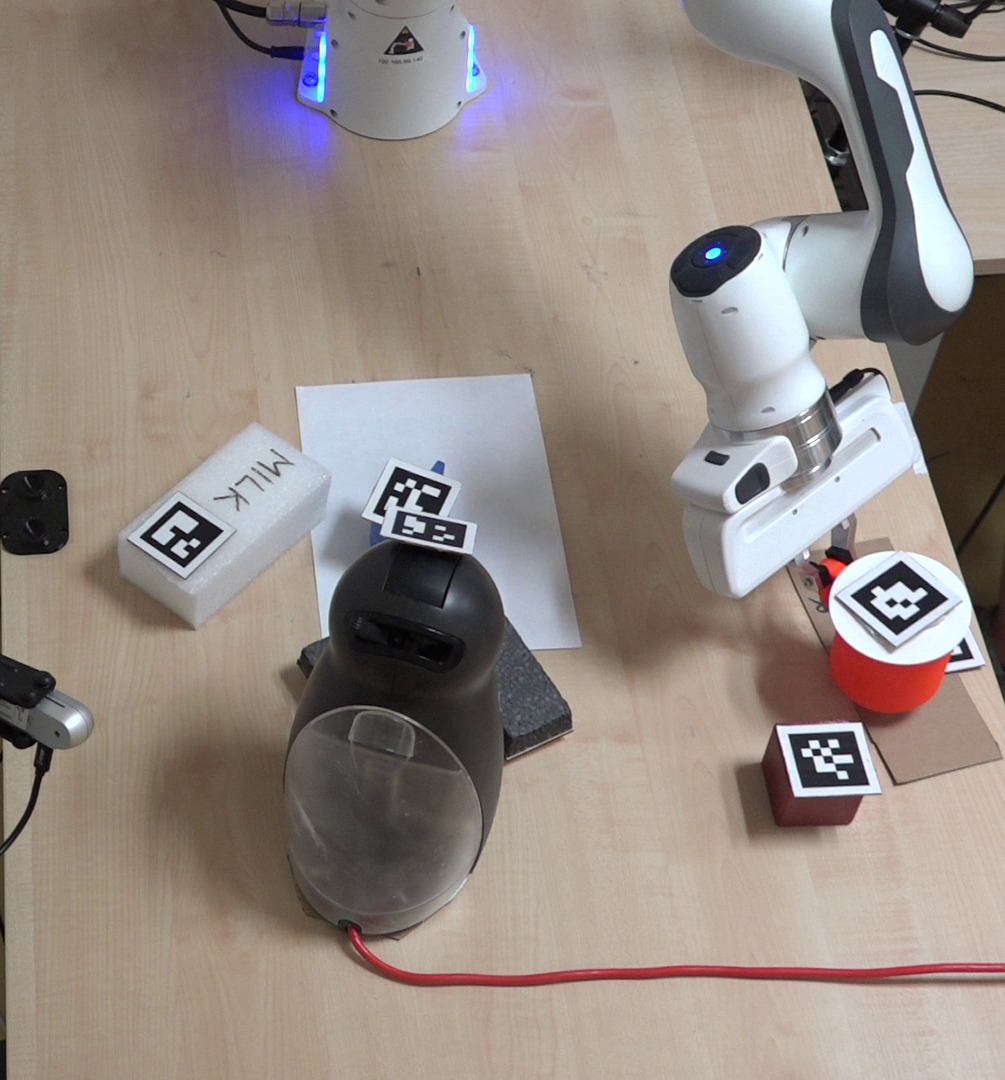}
    \includegraphics[width=0.18\columnwidth]{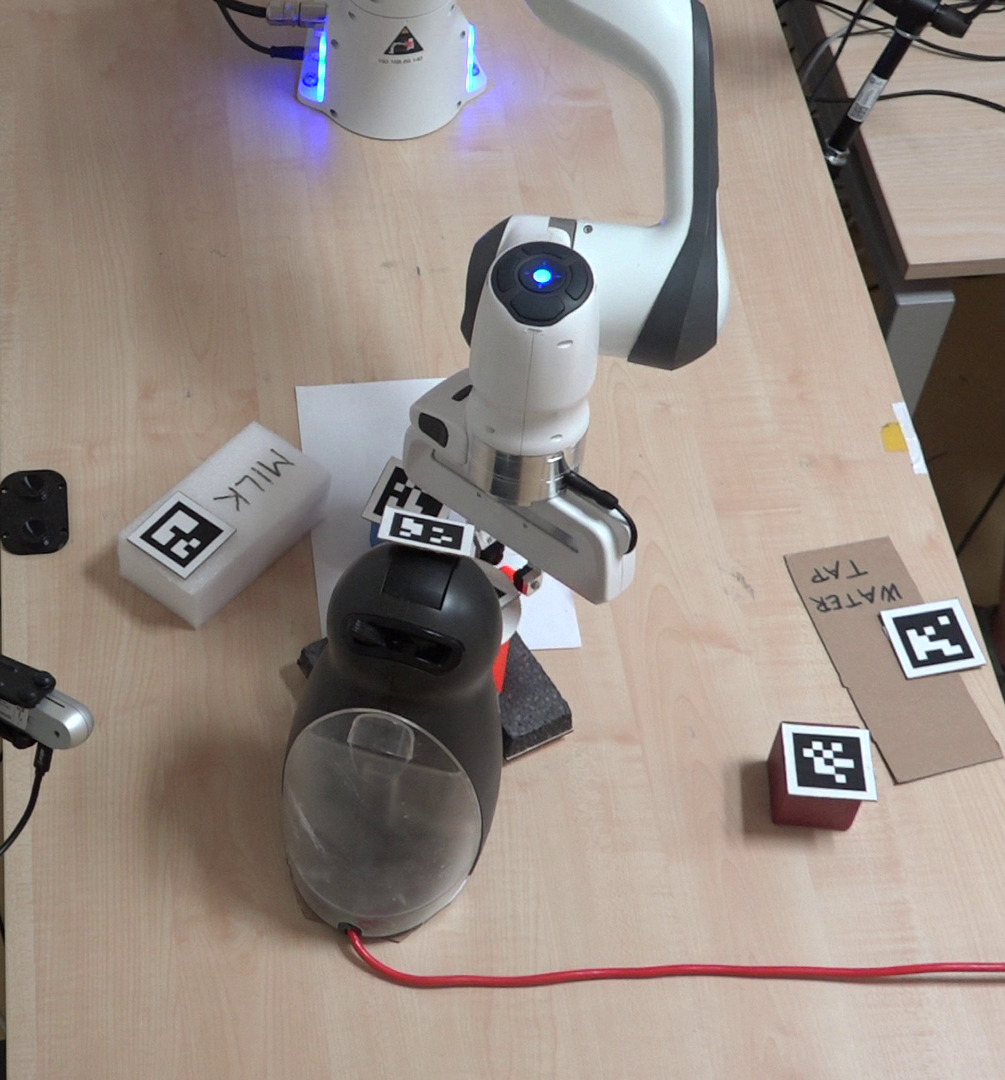}
    \includegraphics[width=0.18\columnwidth]{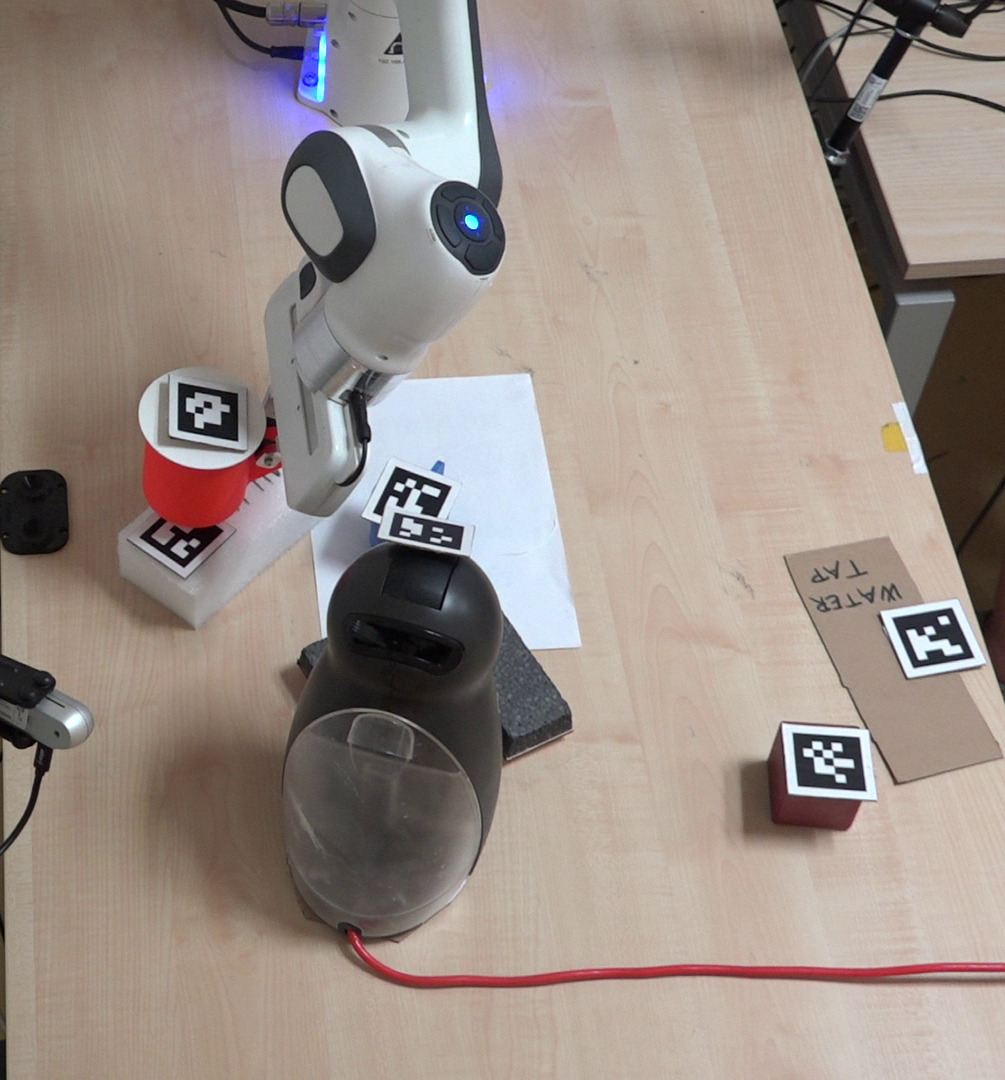}
    \includegraphics[width=0.18\columnwidth]{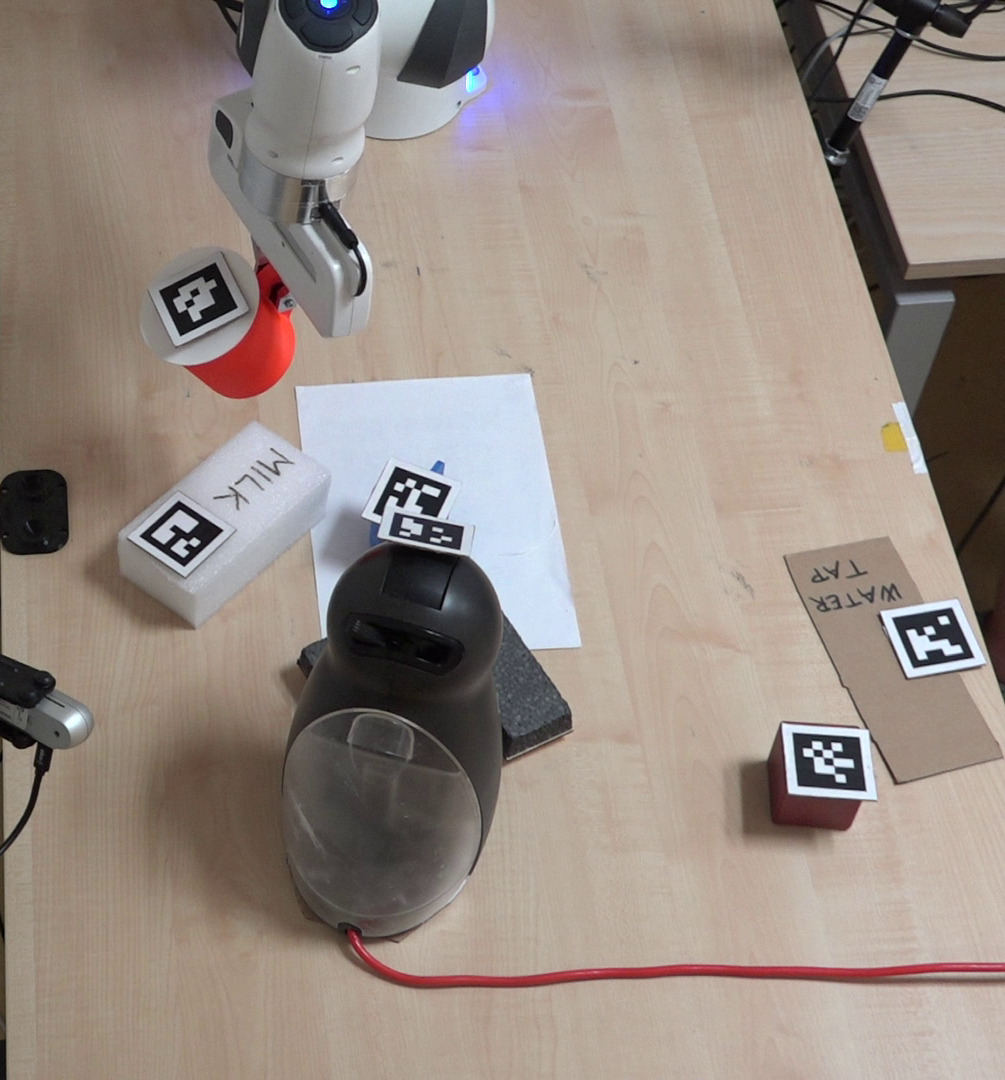}
    \\ \vspace{1mm}
    \includegraphics[width=0.18\columnwidth]{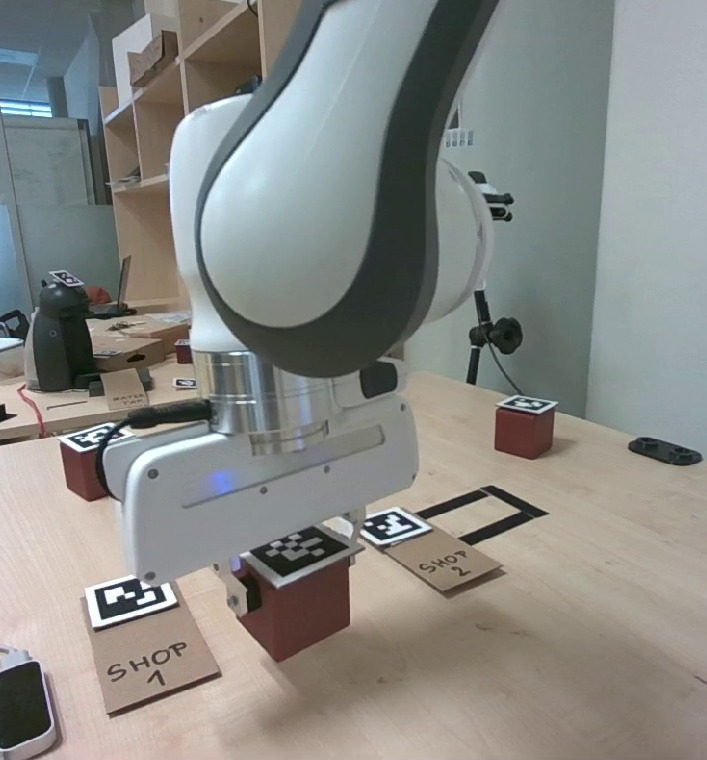}
    \includegraphics[width=0.18\columnwidth]{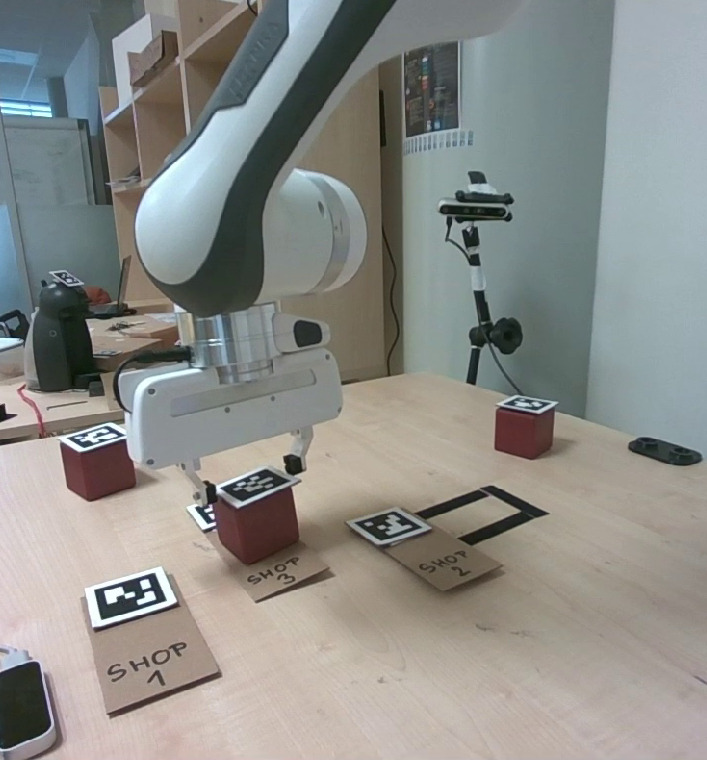}
    \includegraphics[width=0.18\columnwidth]{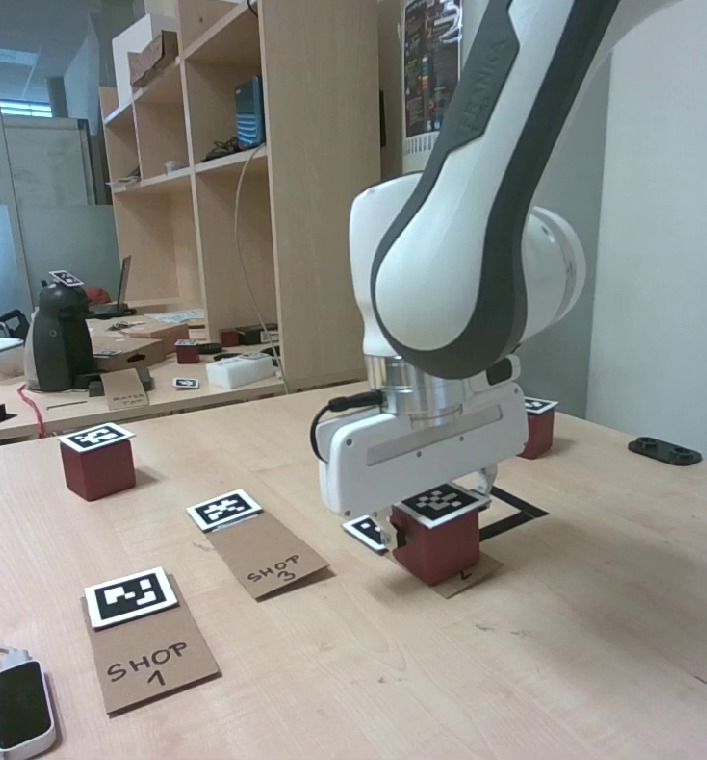}
    \includegraphics[width=0.18\columnwidth]{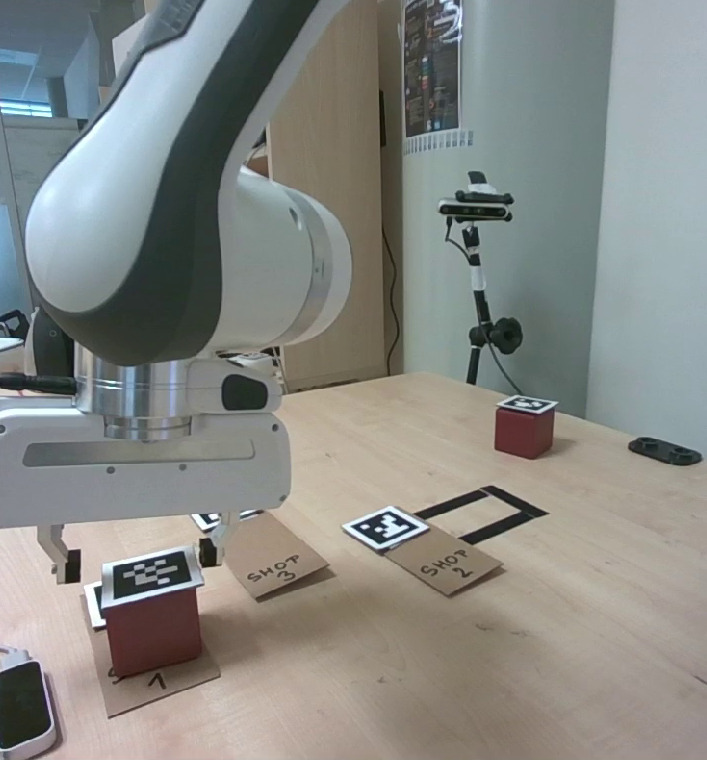}
    \includegraphics[width=0.18\columnwidth]{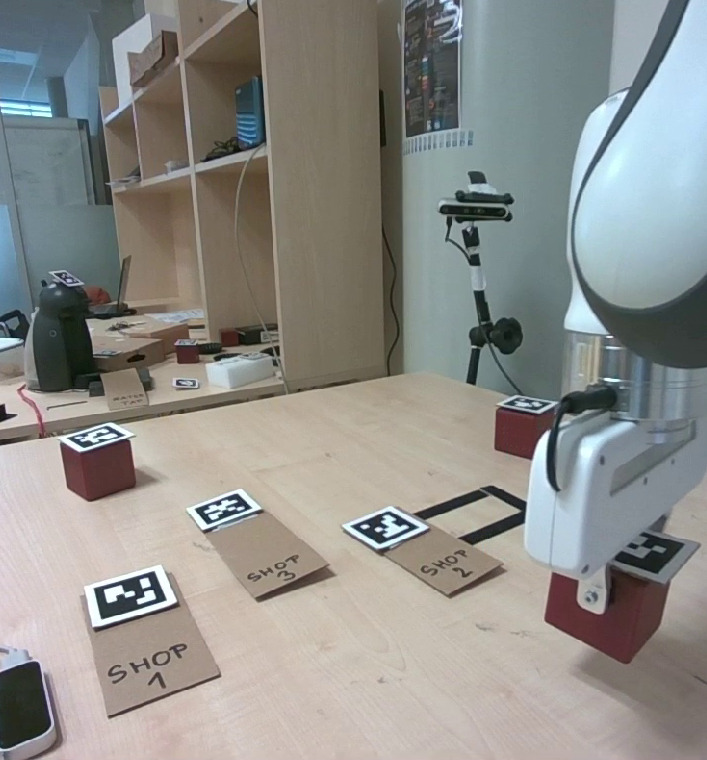}
    \\ \vspace{1mm}
    \includegraphics[width=0.18\columnwidth]{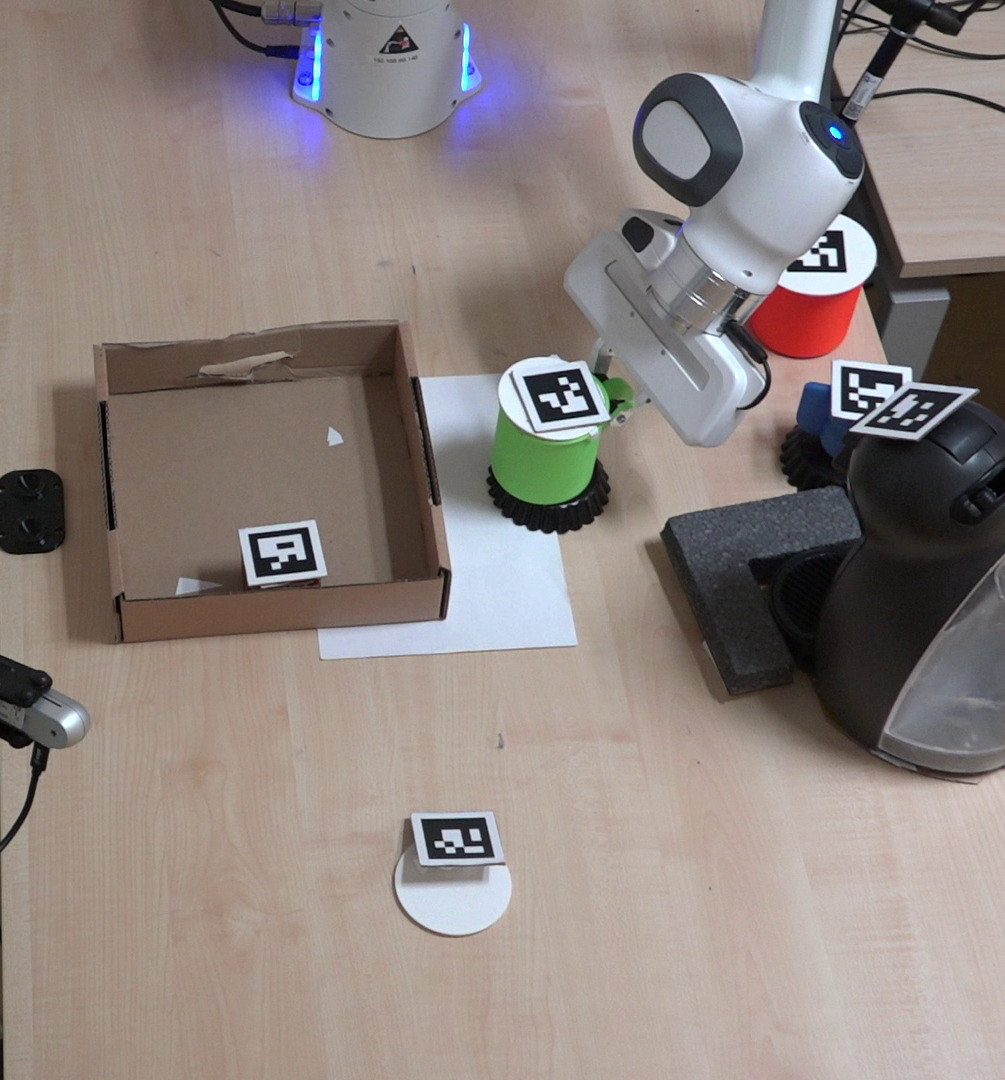}
    \includegraphics[width=0.18\columnwidth]{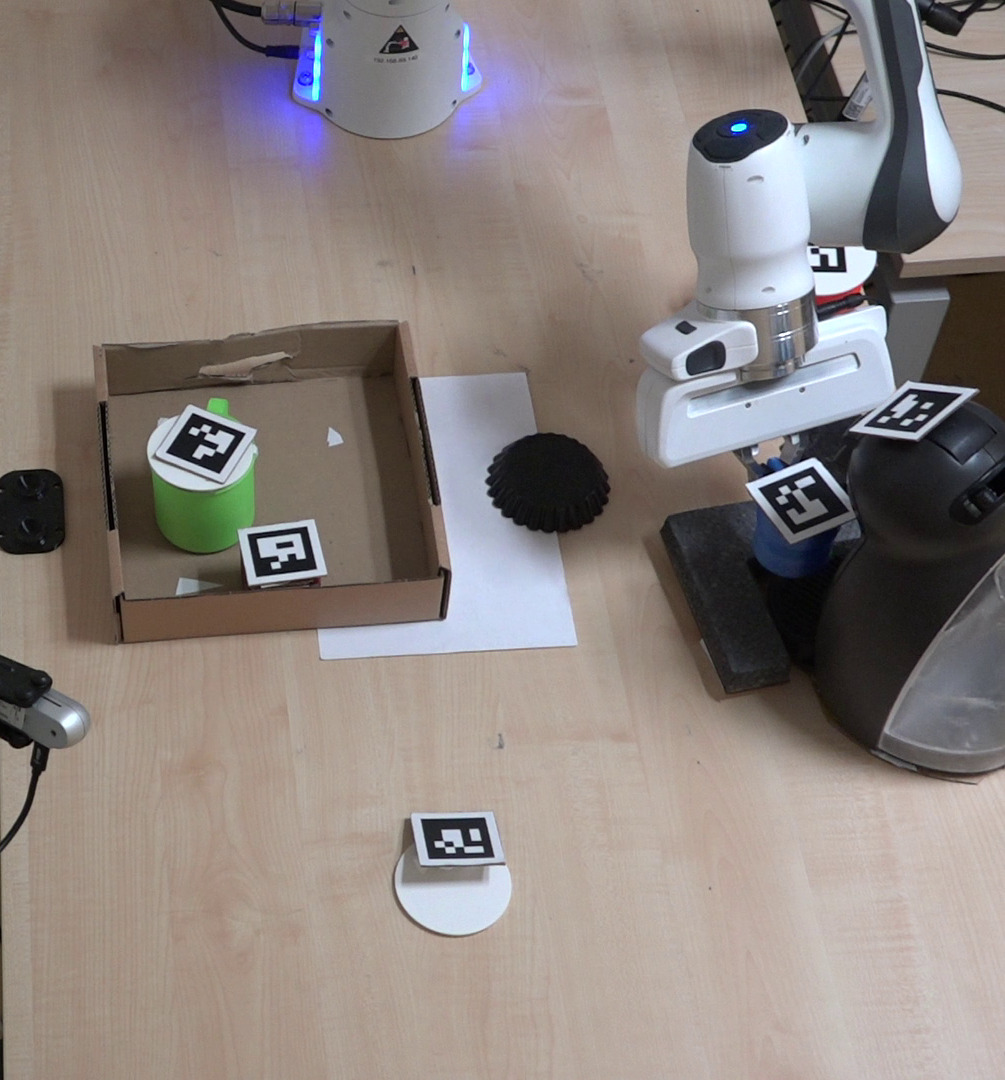}
    \includegraphics[width=0.18\columnwidth]{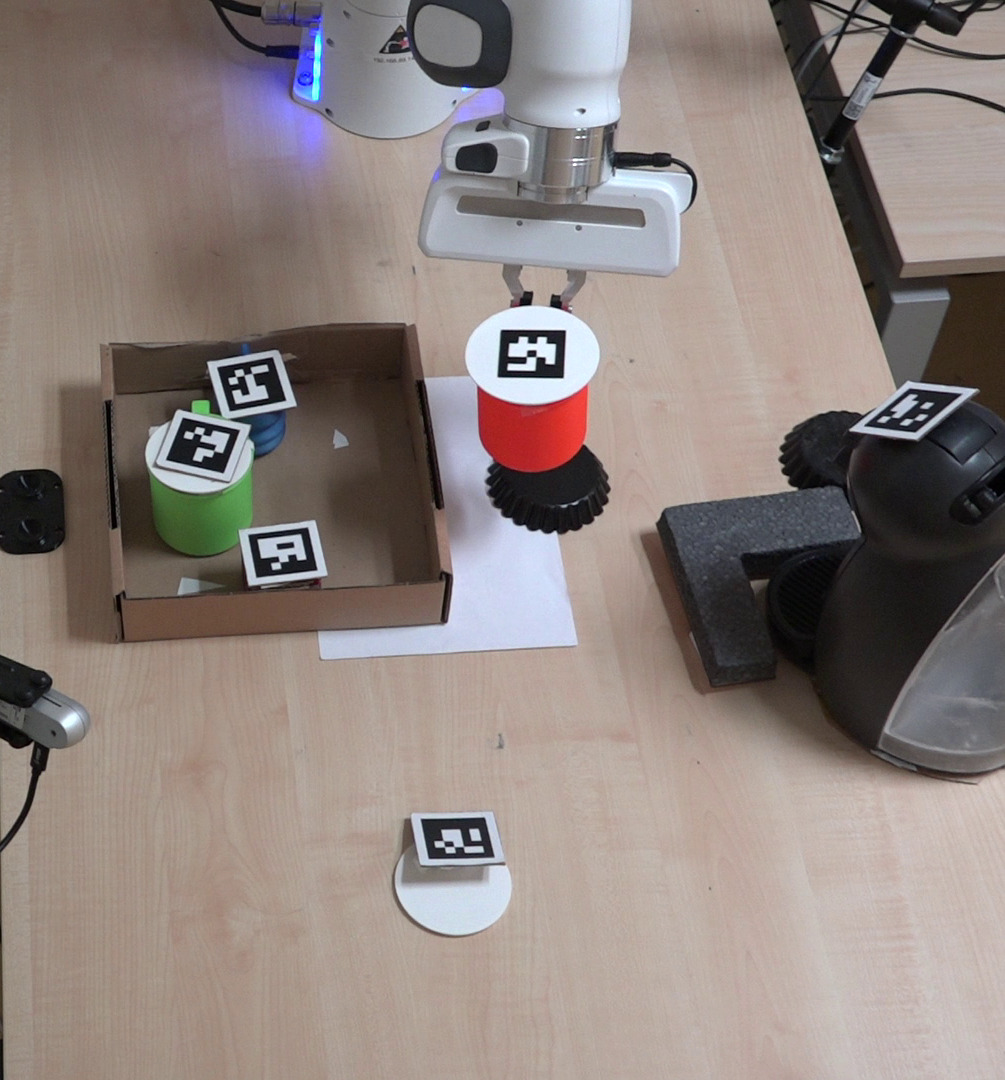}
    \includegraphics[width=0.18\columnwidth]{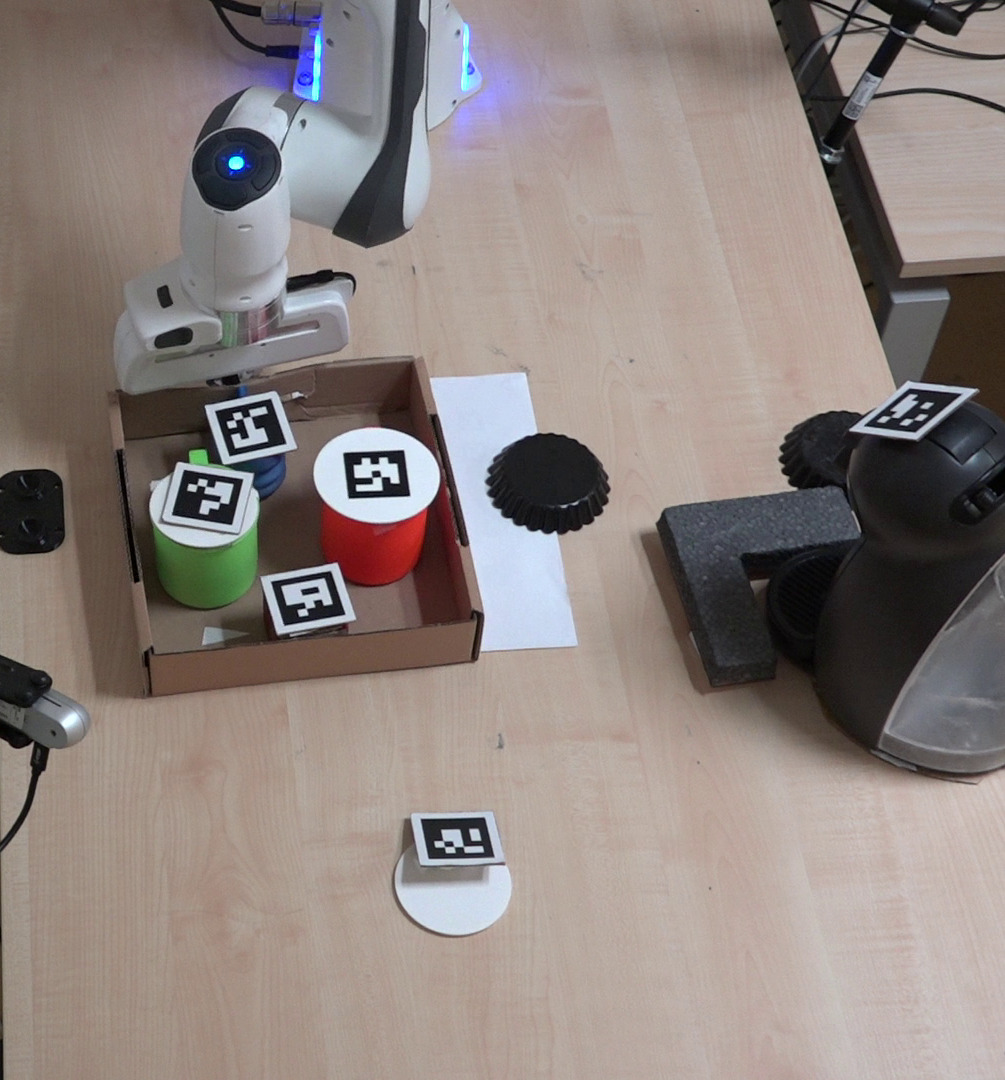}
    \includegraphics[width=0.18\columnwidth]{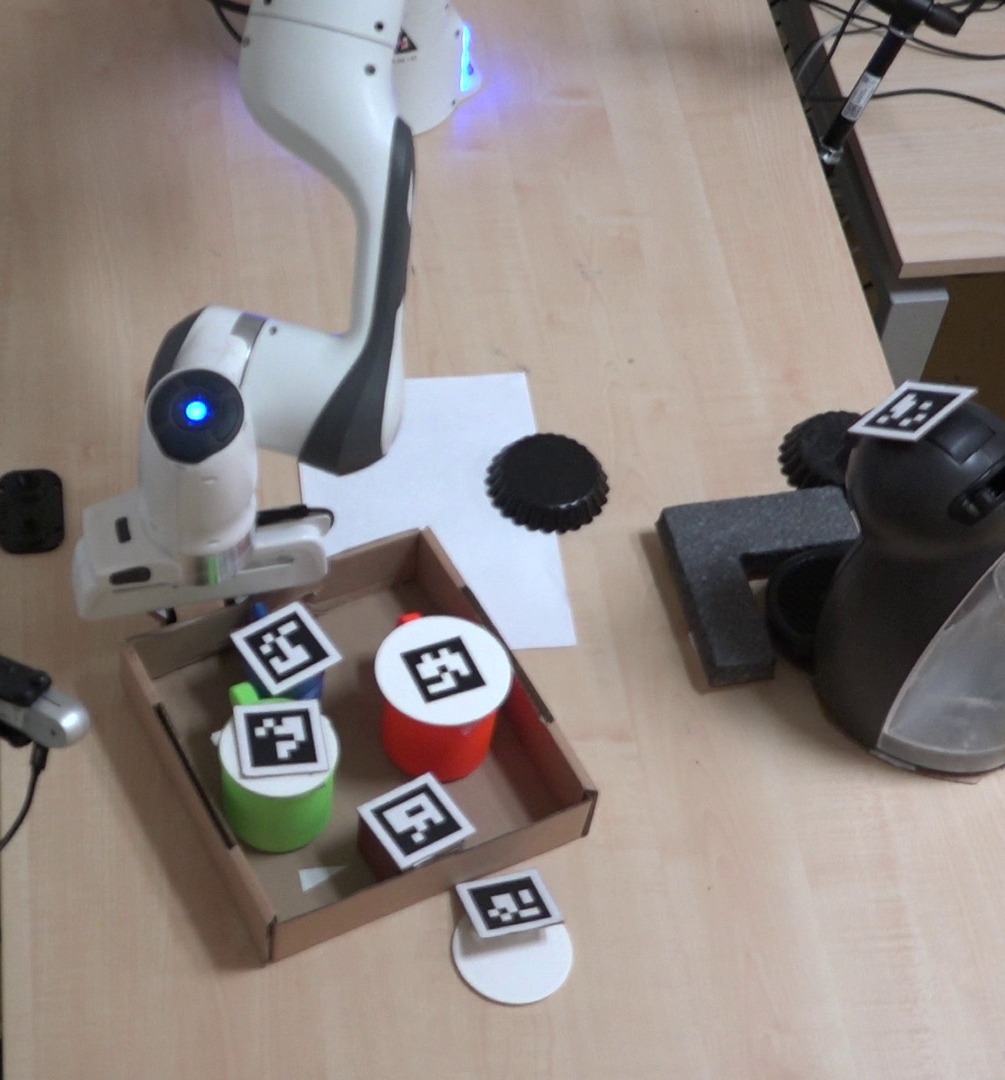}
    \caption{We pick three tasks from our datasets and use the extracted model to drive a robotic execution. The strictness of possible execution orders varies by task, so our execution sampling produces different permissible orders. \textit{Top}: The robot has to move a coffee cup through three strict stations. \textit{Middle}: The robot has to move a cube symbolizing a shopping bag to three shops in an arbitrary order. \textit{Bottom}: The robot has to move three coffee cups to the coffee machine and then to a tray. Finally, the tray is pushed towards a goal location.}
    \label{fig:robot-experiment}
    \vspace{-3mm}
\end{figure}

While the focus of this work is on extracting task dependencies from demonstration data, we still aim to close the loop by integrating it into a robotic application. To this end, we set up a robotic execution system that manipulates different objects by maximizing the probabilities of their poses given the current scene. Our system observes the poses of objects from Aruco markers. For each activation step, the relative pose between the end-effector and the object is recorded and assumed to remain unchanged during robotic manipulation. Our control scheme is driven primarily by maximizing the object's pose probability. If an event requires multiple objects to be manipulated, we constrain the controller to perform only planar pushing motions. 
We reproduce two simple and one complex task from our HANDSOME-COMPLEX dataset. We generate executions from our TPGs. We can sample sensible execution orders from the structures produced by \ourmethod{}, as shown in the video submitted with this manuscript.
However, we also note that the abstract and simple control scheme makes the execution sensitive to object placements.
Here, it would be helpful to introduce more geometric information into the control pipeline~\cite{sundaralingam2026curobov2,kashyap2025single}, or deploy light-weight trajectory cloning (\ie~\cite{vhartz2026unreasonable}) on the manipulation segments encoded in our graph.

\section{Discussion and Conclusion}

In this paper, we introduced \ourmethod{} an approach to extracting Task Precedence Graphs (TPG) from multiple demonstrations of the same task. Different to preceding methods, \ourmethod{} uses not just time or high-level symbolic features to infer mandatory precedences, but is able to directly employ sub-symbolic observations of object poses from the demonstration data. %
To derive the TPG from initial precedence probabilities, we introduce a processing pipeline which slowly dissolves redundant connections. Our evaluation showed that exploiting these sub-symbolic observations is especially useful when one has fewer than a handful of demonstrations, or when one expects the demonstrators to be biased in their execution. In the remainder of our evaluation, we present a study of our method, showing that our new supervision signals must be used in tandem and highlighting the relative benefits of the individual stages of our pipeline. Finally, we demonstrated sampling linear robotic executions from our models.
Despite the general success, we note that \ourmethod{} is not able to capitalize as much on contradictory observations as our baseline. We studied replacing our temporal prior with the contradiction-based one from the baseline, but note that this leads to deteriorated performance in the cases of biased executions again. Here, further study is needed.
In addition, this work neglects one source of supervision which could be quite useful: Prior work~\cite{pardowitz2007incremental,mohseni2015interactive,behrens2019specifying,shah2020interactive} enabled human operators to annotate parts of the model to support its formation. We see potential for doing so with \ourmethod{} as well, by defining a likelihood function for the model given the data and querying human operators about model parts deemed unlikely.

\bibliographystyle{IEEEtran}

\bibliography{bibliography_iros2026}  %

\end{document}